\documentclass[11pt]{article}

\usepackage[preprint]{acl}

\usepackage{times}
\usepackage{latexsym}
\usepackage{url}            
\usepackage{xurl}
\usepackage[T1]{fontenc}
\usepackage[most,skins,theorems]{tcolorbox}

\usepackage[utf8]{inputenc}

\usepackage{microtype}

\usepackage{inconsolata}

\usepackage{graphicx}
\usepackage{wrapfig}
\usepackage[utf8]{inputenc} 
\usepackage[T1]{fontenc}    
\usepackage{hyperref}       
\usepackage{url}            
\usepackage{booktabs}       
\usepackage{amsfonts}       
\usepackage{nicefrac}       
\usepackage{microtype}      
\usepackage{xcolor}         
\PassOptionsToPackage{table}{xcolor}
\usepackage[table]{xcolor}
\usepackage{booktabs}

\definecolor{Gray}{gray}{0.95}
\definecolor{darkspringgreen}{rgb}{0.09, 0.45, 0.27}
\definecolor{darkspringred}{rgb}{0.85, 0.2, 0.2}
\definecolor{americanrose}{rgb}{1.0, 0.01, 0.24}

\usepackage{multirow} 
\usepackage{enumitem}
\usepackage{listings}
\usepackage{caption}
\usepackage{graphicx}
\usepackage{amsthm}

\newtheorem{theorem}{Theorem}
\newtheorem{lemma}[theorem]{Lemma}
\newtheorem{proposition}[theorem]{Proposition}
\newtheorem{observation}[theorem]{Observation}

\lstnewenvironment{normalcode}{
  \lstset{
    language=Python,
    basicstyle=\ttfamily\scriptsize,
    backgroundcolor=\color{white},
    breaklines=true,
    showstringspaces=false,
    frame=none,
    xleftmargin=1em,
    aboveskip=1pt,
    belowskip=1pt,
    breaklines=true,
    xleftmargin=0pt,
  }
}{}
\usepackage{svg}

\lstnewenvironment{diffaddcode}{
  \lstset{
    language=Python,
    basicstyle=\ttfamily\scriptsize,
    backgroundcolor=\color{green!15},
    breaklines=true,
    showstringspaces=false,
    frame=none,
    xleftmargin=1em,
    aboveskip=1pt,
    belowskip=1pt,
    breaklines=true,
    xleftmargin=0pt,
    mathescape=true 
  }
}{}
\usepackage{amsmath}
\lstnewenvironment{diffdelcode}{
  \lstset{
    language=Python,
    basicstyle=\ttfamily\scriptsize,
    backgroundcolor=\color{red!15},
    breaklines=true,
    showstringspaces=false,
    frame=none,
    xleftmargin=1em,
    aboveskip=1pt,
    belowskip=1pt,
    xleftmargin=0pt,
    breaklines=true,
  }
}{}
\DeclareCaptionType{listing}[Listing][List of Listings]
\title{Beyond the Sampled Token: Preserving Candidate Support in RLVR}

\author{
Ruotian Peng$^{1,\dagger}$ \quad Yi Ren$^{2,\dagger}$ \quad Zhouliang Yu$^3$ \quad Weiyang Liu$^3$ \quad Yandong Wen$^{1,*}$ \\[1mm]
$^1$Westlake University~~~$^2$University of British Columbia~~~$^3$The Chinese University of Hong Kong\\[0.5mm]
$^\dagger$Equal contribution~~~~$^*$Corresponding author~~~~\href{https://spherelab.ai/simko/}{\tt {Spherelab.ai/CaSP}} \\
}

\usepackage{minitoc}

\begin{document}

\maketitle

\begin{abstract}
We revisit exploration collapse in reinforcement learning with verifiable rewards (RLVR), from the perspective of the \emph{candidate distribution} for next-token prediction. We formally show that as probability concentrates on the top-$1$ candidate, the expected number of distinct responses collapses to one regardless of the sampling budget $K$. This theoretical implication is further verified by our empirical tracking of top-$N$ candidate probabilities during training, where the top-$1$ candidate progressively dominates while plausible alternatives are suppressed. These findings suggest a key desideratum for effective exploration: \emph{preserving non-negligible probability mass on the top-$N$ candidates}. To this end, we propose Candidate-aware Support Preservation (CaSP), with two complementary designs. Specifically, CaSP redistributes positive gradients among top-$N$ candidates for correct responses, and applies a stronger penalty to the top-$1$ candidate for incorrect responses. Unlike many exploration-oriented methods that improve pass@$K$ at the cost of pass@1, CaSP improves pass@$K$ across the full $K$ spectrum. These gains generalize to 6 math, 2 logical-reasoning, and 2 coding benchmarks, and scales to 32B-parameter models and sampling budgets up to $K=1024$, positioning it as a principled, candidate-level approach for RLVR exploration.
\end{abstract}

\section{Introduction}
Reinforcement learning with verifiable rewards (RLVR) offers a simple recipe for improving LLM reasoning: the model generates responses, and updates itself by rewarding correct ones and penalizing incorrect ones \cite{shao2024deepseekmath, schulman2017proximal, hu2025reinforce++}. This coupled process induces a bias towards exploitation over exploration, whereby the model collapses to a narrow set of responses \cite{liang2025beyond, he2025rewarding}. Such an effect is evident in improved pass@$1$, which measures how often a single response is correct, but degraded pass@$K$ ($K$>1), which measures whether the model can find correct solutions with many attempts \cite{yue2025does,wu2025invisible}. Without sufficient exploration, the model struggles with more challenging scenarios, ultimately capping its potential for further reasoning improvement \cite{chen2025pass, song2025outcome}.
\begin{figure*}[t]
    \centering
    \includegraphics[width=1\textwidth ]{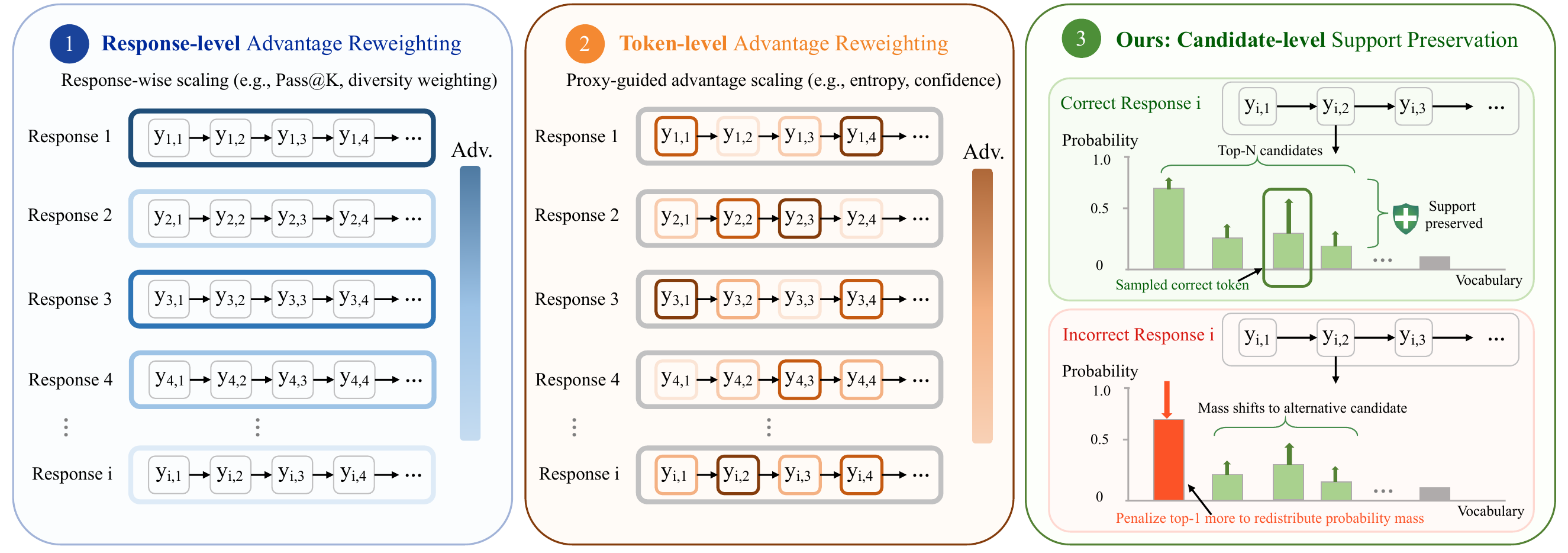}
  \caption{Overview of exploration-oriented methods at different intervention granularities. Existing approaches mainly operate at the response level or token level, whereas CaSP operates at the candidate level by explicitly preserving probability mass among top-$N$ plausible next-token candidates.}
    \label{fig:simko_method}
\end{figure*}

To address this exploration deficit, existing approaches modify the optimization signal at different granularities, with conceptual comparison in Figure \ref{fig:simko_method}. \emph{Response-level} advantage shaping reweights entire responses, either through pass@$K$  objectives \citep{chen2025pass, walder2025pass} or through response rarity, diversity, and correctness signals \citep{liu2026past, gai2025differential, zhu2025surprising}. These methods improve selection among sampled responses but do not explicitly encourage exploring new ones. \emph{Token-level} advantage shaping operates at each decoding step, typically through entropy-based bonuses computed from the next-token distribution \citep{cheng2025reasoning, cui2025entropy, jiang2025rethinking}. However, entropy is too coarse to distinguish which candidates deserve probability mass, and may push it onto unrelated ones. Sitting between these levels, \emph{segment-level} approaches shape advantages over contiguous token spans \citep{chen2025beyond}. \emph{Hybrid} approaches combine response- and token-level signals, through curiosity-driven or outcome-conditioned rewards \citep{dai2025cde, song2025outcome}. Overall, existing approaches either insufficiently expand the response space or expand it without control, falling short of effective  exploration.

We take an alternative perspective grounded in next-token prediction. At each decoding step, an LLM outputs a candidate distribution over its vocabulary\footnote{The \emph{vocabulary} is the set of all possible \emph{candidates}; one \emph{candidate} is sampled and becomes the \emph{token} in a \emph{response}.}, offering a fine-grained view of its exploration behavior. How probability is distributed across candidates determines whether the model explores multiple reasoning paths or collapses into a single deterministic response \cite{deng2025token,zhu2025surprising}, as formally stated in Proposition~\ref{prop:local_diversity}. Yet capturing the full distributioni is computationally prohibitive since modern vocabularies exceed 100K candidates. This likely explains why prior work favored scalar measures like entropy.

\begin{figure*}[!ht] 
    \centering
    \includegraphics[width=1\linewidth]{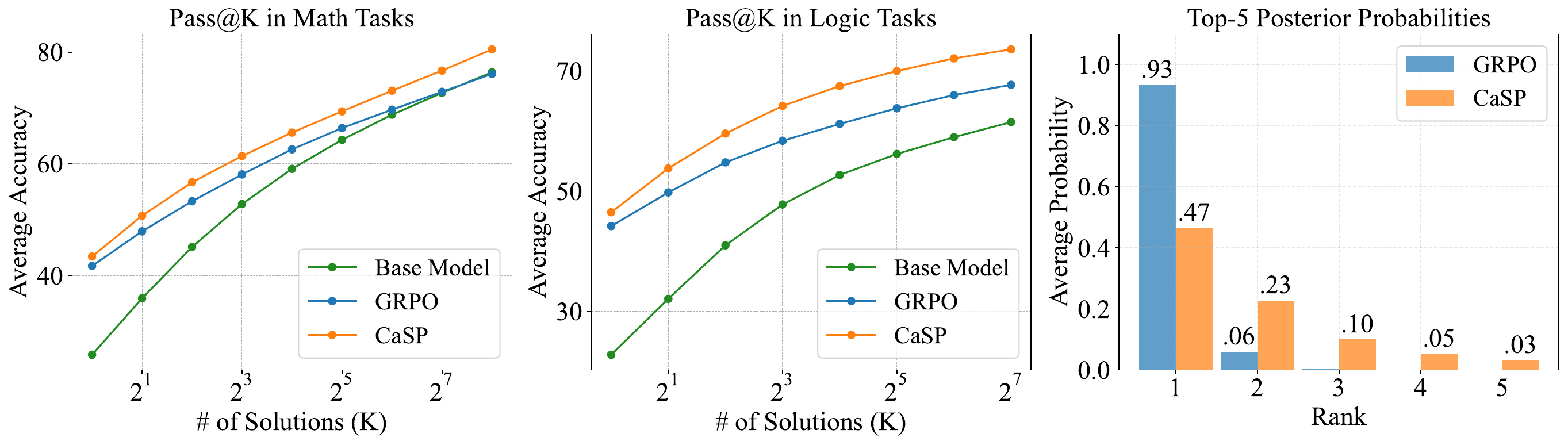}
    \caption{\textbf{Left/middle}: pass@$K$ of CaSP vs.\ GRPO on math (AIME24/25, AMC, MATH500, Minerva, OlympiadBench) and logic (SynLogic, BBH) benchmarks; CaSP improves across the $K$ spectrum. \textbf{Right}: average probability of the rank-$n$ candidate; CaSP's distribution is less concentrated than GRPO's.}
    \label{fig:teaser}
\end{figure*}

We sidestep this constraint by examining candidate distributions empirically. These distributions are highly skewed (Figure~\ref{fig:teaser}), with only a few candidates carrying non-negligible probability mass, so the top-$N$ candidates serve as a tractable yet faithful approximation of the full distribution. During RLVR fine-tuning, we observe that probability mass progressively concentrates on the top-$1$ candidate while plausible alternatives are suppressed. This pattern matches the Proposition~\ref{prop:local_diversity} and directly explains the pass@$K$ degradation. These findings point to a clear desideratum for effective exploration: \emph{preserving non-negligible probability mass on the top-$N$ candidates}.

To achieve this desideratum, we propose Candidate-aware Support Preservation (CaSP), with two complementary designs. For correct responses, CaSP redistributes positive gradients among the top-$N$ candidates, smoothing reinforcement across plausible alternatives rather than concentrating it on the sampled token. For incorrect responses, CaSP applies a stronger penalty to the top-$1$ candidate to counteract the squeeze effect, redirecting probability mass to alternatives. These updates prove especially effective when applied to high-entropy tokens in the reasoning path.

We evaluate CaSP on multiple LLM backbones and a diverse set of benchmarks, including 6 math, 2 logical reasoning, and 2 coding tasks. Unlike GRPO (which improves pass@$1$ but struggles at pass@$K$) and exploration-oriented baselines (which improve pass@$K$ but sacrifice pass@$1$), CaSP achieves the best of both worlds, improving pass@$K$ across the full spectrum (Figure~\ref{fig:teaser}). Moreover, these gains scale to models up to 32B parameters and sampling budgets up to $K=1024$. In summary, our work makes three contributions:

\begin{itemize}[leftmargin=*,nosep]
\setlength\itemsep{0.43em}
\item We introduce a candidate-distribution perspective on RLVR exploration, supported by both formal analysis and empirical verification.
\item We propose CaSP, a principled candidate-level method that encourages exploration by preserving top-$N$ candidate supports.
\item CaSP's gains generalize to multiple LLM backbones and diverse reasoning benchmarks, scaling to 32B models and $K=1024$ sampling budgets.
\end{itemize}

\section{Background and Preliminaries}
\label{sec2}

RLVR trains an LLM using rewards from a programmatic verifier (\emph{e.g.,} a math grader or code checker). Group relative policy optimization (GRPO) \cite{shao2024deepseekmath}, is a widely used RLVR method tailored for LLM post-training. Given a question $x$, the model generates $G$ different responses $\{y_i\}_{i=1}^G$ and updates its parameters $\theta$ as:
\begin{equation*}
\begin{aligned}
& \mathcal{J}_{\text{GRPO}}(\theta; \gamma_{i,t}) = \frac{1}{G} \sum_{i=1}^{G} \frac{1}{|y_i|} \sum_{t=1}^{|y_i|}  \Big[ \\
& \quad \min \big( \gamma_{i,t} A_i, \mathsf{clip}_\epsilon(\gamma_{i,t}) A_i \big) - \beta \, \mathbb{D}_\text{KL}(\pi_\theta \Vert \pi_\text{ref}) \Big]
\end{aligned}
\end{equation*}
where $\mathsf{clip}_\epsilon(\gamma)$ truncates the ratio $\gamma$ to $[1\!-\!\epsilon, 1\!+\!\epsilon]$, and $\gamma_{i,t} = {\pi_\theta(y_{i,t}|s_{i,t})}/{\pi_{\text{old}}(y_{i,t}|s_{i,t})}$ is the likelihood ratio between the current policy $\pi_\theta$ and the previous-iteration policy $\pi_{\text{old}}$ at the $t$-th token of response $y_i$, with $s_{i,t} = (x, y_{i,<t})$ denoting the decoding state. The KL term uses $\pi_{\text{ref}}$, the frozen pre-trained reference. The advantage $A_i = (r_i - \mathrm{mean}(\{r_j\}_{j=1}^G))/\mathrm{std}(\{r_j\}_{j=1}^G)$ is computed by normalizing the verifier reward $r_i\in \{0, 1\}$ of response $y_i$ across the group of $G$ rollouts.

GRPO can be analyzed in gradient space under the learning dynamics framework of \cite{ren2024learning}. The dominant contribution to parameter updates comes from $\nabla_\theta A_i\gamma_{i,t}$. Here, the advantage $A_i$ can be viewed as a response-level adaptive learning rate that scales gradients from different responses $y_i$. Ignoring the KL and clipping terms (both primarily introduced for stability), the main optimization direction is:
\begin{equation}
\label{eq:gradient_ratio}
\nabla_\theta A_i \gamma_{i,t} = A_i \cdot \mathrm{sg}(\gamma_{i,t}) \cdot \nabla_\theta\log \pi_{\theta}(y_{i,t}|s_{i,t}),
\end{equation}
where $\mathrm{sg}(\cdot)$ is the stop-gradient operator. CaSP builds on this gradient form to redistribute mass across the candidate distribution.

\section{A Candidate-Level Analysis of Why RLVR Concentrates Probability}
\label{section3}
In this section, we show that (i) both positive and negative RLVR updates push probability mass toward the rank-$1$ candidate, (ii) this candidate-level concentration leads to low-diversity responses, and (iii) empirical tracking rank-wise log-probabilities across RLVR methods reproduces the predicted concentration. These results highlight the importance of preserving supports over the top-$N$ candidates, which we take as a natural desideratum for effective exploration.

\subsection{RLVR Updates Reinforce Concentration}
We now show why on-policy RLVR updates natually push candidate distribution toward concentration, considering positive and negative samples separately. Let $\mathcal{V}$ denote the vocabulary; for any candidate $v \in \mathcal{V}$ at state $s_{i,t}$, let $z_v$ and $p_v = \pi_\theta(v|s_{i,t})$ be its logit and probability.

\begin{observation}[Positive updates amplify dominant candidates]
\label{obs:preferential_exposure}
At state $s_{i,t}$, GRPO samples tokens from $\pi_\theta(\cdot | s_{i,t})$, so the expected positive-advantage reinforcement of each candidate is proportional to its current probability $p_v$.
\end{observation}

Since the rank-$1$ candidate is sampled most frequently and therefore reinforced most by positive advantages. The training process widens the probability gap between the rank-$1$ candidate and lower-ranked alternatives, driving the next-token distribution toward over-concentration.

Prior work \cite{zhu2025surprising} suggests that negative samples could counteract concentration by suppressing incorrect token and redistributing probability mass to alternatives \cite{zhu2025surprising}. However, this redistribution flows disproportionately to high-probability candidates, known as the squeeze effect \cite{ren2024learning}.

\begin{lemma}[Squeeze effect of negative updates]
\label{lem:squeeze_effect}
For the unclipped GRPO term $A_i \gamma_{i,t}$ with $A_i < 0$, the gradient decreases the sampled token's probability $p_{y_{i,t}}$. For any alternative $v \neq y_{i,t}$, the softmax coupling gives:
\begin{equation*}
\label{eq:squeeze_non_sampled}
    \frac{\partial p_{y_{i,t}}}{\partial z_v} = -p_{y_{i,t}} p_v.
\end{equation*}
The negative update therefore raises $z_v$ in proportion to $p_{y_{i,t}} p_v$. If the penalized token $y_{i,t}$ is not the rank-$1$ candidate, the rank-$1$ candidate absorbs the largest increase. (See Appendix~\ref{app:proof_squeeze_effect}.)
\end{lemma}


\begin{figure*}[t]
  \centering
  \setlength{\abovecaptionskip}{4pt}
  \setlength{\belowcaptionskip}{-7pt}
  \includegraphics[width=1\textwidth,]{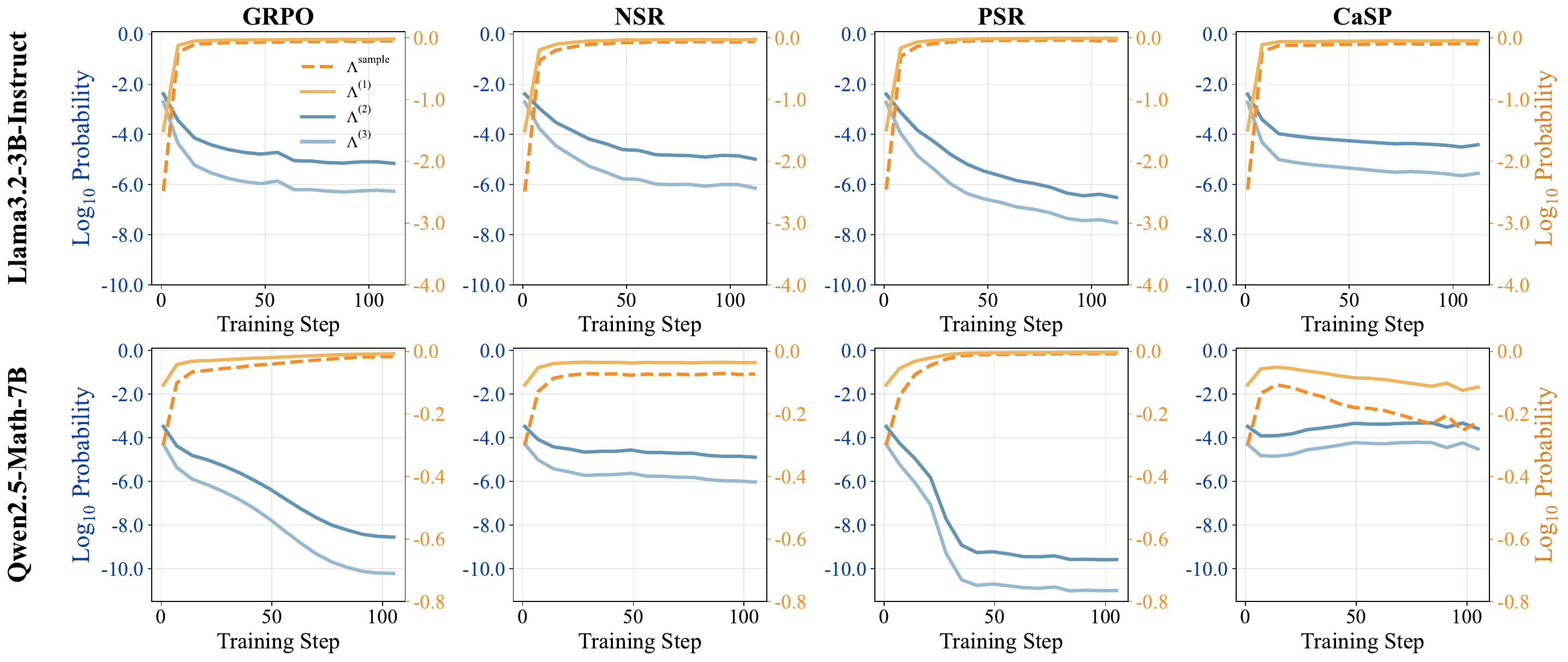}
\caption{Training dynamics of the sampled-token log-probability $\Lambda^{\mathrm{sample}}$ and rank-$n$ candidate log-probabilities $\Lambda^{(n)}$ across RLVR variants: GRPO, PSR (training with only positive samples), NSR (training with only negative samples), and CaSP. Following the setup of \cite{zhu2025surprising}, we train Llama3.2-3B-Instruct on a mixture of GSM8K and MATH (Level 1) and Qwen2.5-Math-7B on the MATH dataset.}
  \label{fig:analysis_onpolicy}
\end{figure*}
\subsection{From Candidate to Response Collapse}
Positive and negative updates push mass to the rank-$1$ candidate. We now connect this concentration to pass@$K$ through a simple sampling fact.

\begin{proposition}[Local concentration limits rollout diversity]
\label{prop:local_diversity}
At state $s_{i,t}$ with distribution $p$ over $\mathcal{V}$, consider $K$ independent rollouts that reach this state and sample their next tokens $\{\tilde{y}_{1,t},\ldots,\tilde{y}_{K,t}\}$. The expected number of distinct next tokens is
\begin{equation*}
  \mathbb{E}
  \left[
  \left|
  \{\tilde{y}_{1,t},\ldots,\tilde{y}_{K,t}\}
  \right|
  \right]
  =
  \sum_{v\in\mathcal{V}}
  \left(1-(1-p_v)^K\right).
\end{equation*}
If the rank-$1$ candidate probability $p_{v^{(1)}}$ converges to $1$, this expectation converges to $1$ for any fixed sampling budget $K$.
\end{proposition}

Proposition~\ref{prop:local_diversity} connects candidate-level concentration to rollout-level exploration. When the distribution at a branching state collapses to one dominant candidate, many independent rollouts choose the same next token and follow similar reasoning paths. Local loss of token diversity therefore leads to homogeneous trajectories, reducing the marginal benefit of increasing the sampling budget $K$.

\subsection{Empirical Verification}
We empirically verify the concentration by tracking rank-wise log-probabilities during training. For rank-$n$, we measure the average log-probability of the rank-$n$ candidate $v^{(n)}_{i,t}$ at state $s_{i,t}$:
\begin{equation}
\label{eq:logp_topn}
    \Lambda^{(n)}
    :=
    \frac{1}{G}
    \sum_{i=1}^{G}
    \frac{1}{|y_i|}
    \sum_{t=1}^{|y_i|}
    \log \pi_{\theta}(v_{i,t}^{(n)} \mid s_{i,t}).
\end{equation}
We similarly define $\Lambda^{\mathrm{sample}}$ by replacing $v_{i,t}^{(n)}$ with the sampled token $y_{i,t}$. Figure~\ref{fig:analysis_onpolicy} shows how these metrics evolve under different methods.

Under GRPO, $\Lambda^{\mathrm{sample}}$ gradually moves toward $\Lambda^{(1)}$, indicating that sampled tokens increasingly coincide with the rank-$1$ candidate. PSR further amplifies this movement, matching the rank-$1$ amplification predicted by Observation~\ref{obs:preferential_exposure}. In contrast, NSR slows the movement but does not eliminate it, consistent with the squeeze effect described in Lemma~\ref{lem:squeeze_effect}. Together, these patterns show that RLVR updates progressively enlarge the gap between $\Lambda^{(1)}$ and lower-ranked $\Lambda^{(n)}$, thereby concentrating samples around the rank-$1$ candidate and reducing rollout-level diversity. These findings motivate the desideratum: \emph{preserving non-negligible probability mass on the top-$N$ candidates.}

\begin{figure*}[t]
    \centering
    \setlength{\abovecaptionskip}{3pt}
    \setlength{\belowcaptionskip}{-6pt}
    \includegraphics[width=1\textwidth ]{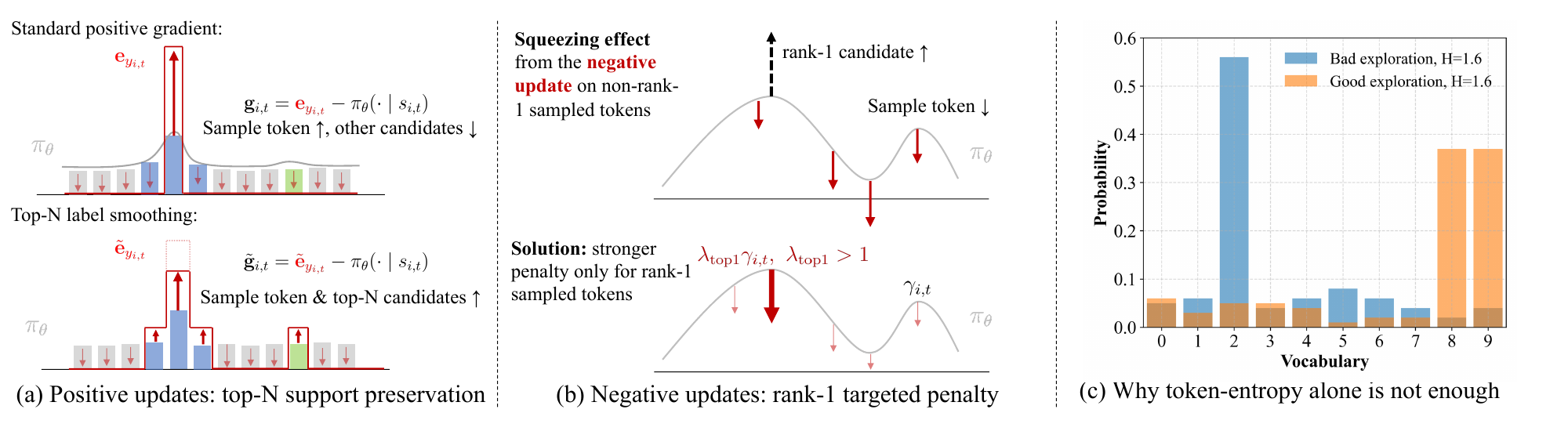}
    \caption{Intuition of CaSP. (a) For positive samples, we redistribute the probability mass from the top-$1$ candidate to the top-$N$ candidates, mitigating over-concentration. (b) For negative samples, we apply a stronger penalty only
  when the sampled token is the current rank-$1$ candidate, avoiding the squeeze effect.  (c) An example of two distributions with identical entropy but distinct probability distributions.}
    \label{fig:methods}
\end{figure*}

\section{Candidate-aware Support Preservation}
\label{sec:casp}

Section~\ref{section3} shows that both positive and negative RLVR updates drive candidate probabilities toward over-concentration, motivating the desideratum of preserving non-negligible probability mass on the top-$N$ candidates. To achieve this desideratum, we propose CaSP, comprising two complementary designs: (i) top-$N$ label smoothing for positive gradients, and (ii) stronger rank-$1$ penalties for negative gradients. Both components are applied only to high-entropy tokens, which often serve as ``forking'' points \cite{deng2025token,wang2025beyond} (selected when their entropy exceeds the $q$-quantile threshold $\tau_q$ within the response).

\subsection{Positive Updates on Top-\texorpdfstring{$\boldsymbol{N}$}{N} Candidates}

Observation~\ref{obs:preferential_exposure} indicates that positive updates tend to amplify already dominant candidates. We make this effect explicit at the gradient level. For the sampled token $y_{i,t}$, let $\mathbf{e}_{y_{i,t}}$ be its one-hot target. The standard log-likelihood gradient is
\begin{equation}
    \mathbf{g}_{i,t}
    = \frac{\partial \log \pi_{\theta}(y_{i,t} | s_{i,t})}{\partial \mathbf{z}_{i,t}}
    = \mathbf{e}_{y_{i,t}} \!- \pi_\theta(\cdot | s_{i,t}).
    \label{eq:G}
\end{equation}
As shown in Figure~\ref{fig:methods}-(a), positive updates increase the sampled token's probability while suppressing alternatives, which sharpens the distribution when the sampled token is often rank-$1$ candidate.

To counter this effect, CaSP reallocates part of the positive update from the sampled token to other plausible candidates via a localized variant of label smoothing \cite{muller2019does}. Unlike vanilla label smoothing, which spreads mass over the full vocabulary and can allocate probability to ungrammatical or irrelevant candidates, our variant smooths only over the model's top-$N$ candidates:
\begin{equation}
\begin{aligned}
    \tilde{\mathbf{e}}_{y_{i,t}}
    &=
    (1-\alpha)\mathbf{e}_{y_{i,t}}
    +
    \frac{\alpha}{N}
    \sum_{n=1}^N
    \mathbf{e}_{v_{i,t}^{(n)}}, \\
    \tilde{\mathbf{g}}_{i,t}
    &=
    \tilde{\mathbf{e}}_{y_{i,t}} - \pi_\theta(\cdot | s_{i,t}),
    \label{eq:G_for_positive}
\end{aligned}
\end{equation}
where $\alpha\in[0,1]$ controls the smoothing strength. This definition is unchanged when the sampled token itself belongs to the top-$N$ set, ensuring consistent treatment of the sampled token. Equivalently, this gradient can be implemented by replacing the original ratio
$\gamma_{i,t}$ with
\begin{equation*}
    \gamma^+_{i,t}
    =
    (1\!-\!\alpha)\gamma_{i,t}
    +
    \frac{\alpha}{N} \!
    \sum_{n=1}^N
    \mathrm{sg}\! \left(\frac{\gamma_{i,t}}{\gamma^{(n)}_{i,t}}\right)
    \gamma^{(n)}_{i,t},
    \label{eq:gamma_pos}
\end{equation*}
where
$\gamma^{(n)}_{i,t} = {\pi_\theta(v_{i,t}^{(n)} | s_{i,t})}/{\pi_{\theta_{\mathrm{old}}}(v_{i,t}^{(n)} | s_{i,t})}$ is the ratio evaluated on the rank-$n$ candidate. The stop-gradient term preserves the importance-sampling coefficient of the sampled token; see Appendix~\ref{section:app:proof} for details. Optimizing with $\gamma^+_{i,t}$ redistributes positive gradients across plausible top-$N$ candidates, forming a plateau rather than a sharp peak, as illustrated in Figure~\ref{fig:methods}-(a).

\subsection{Negative Updates on Top-\texorpdfstring{$\boldsymbol{1}$}{1} Candidate}
\label{sec:negative_updates}

Lemma~\ref{lem:squeeze_effect} shows that negative updates depend on which token is penalized: penalizing rank-$1$ can flatten the distribution, while penalizing lower-ranked tokens can squeeze mass toward rank-$1$. Thus, uniformly amplifying negative terms \citep{zhu2025surprising} also strengthens non-rank-$1$ penalties, which sharpen the distribution via the squeeze effect, as shown in the upper panel of Figure~\ref{fig:methods}-(b).

CaSP instead applies stronger negative updates only when the sampled token is the current rank-$1$ candidate. For tokens with $A_i<0$, we replace the original ratio with
\begin{equation}
\label{eq:gamma_neg}
    \gamma^-_{i,t}
    =
    \begin{cases}
    \lambda_{\mathrm{top1}}\gamma_{i,t}, & y_{i,t}=v_{i,t}^{(1)},\\
    \gamma_{i,t}, & y_{i,t}\neq v_{i,t}^{(1)},
    \end{cases}
\end{equation}
where $\lambda_{\mathrm{top1}}>1$ controls the strength of the rank-$1$ penalty. This penalizes overconfident incorrect predictions strongly while avoiding the squeeze effect on lower-ranked candidates, as shown in the lower panel of Figure~\ref{fig:methods}-(b).

\subsection{A Didactic Toy Example}
\begin{figure*}[!ht]
  \centering
  \setlength{\abovecaptionskip}{4pt}
  \setlength{\belowcaptionskip}{-7pt}
  \includegraphics[width=1\textwidth]{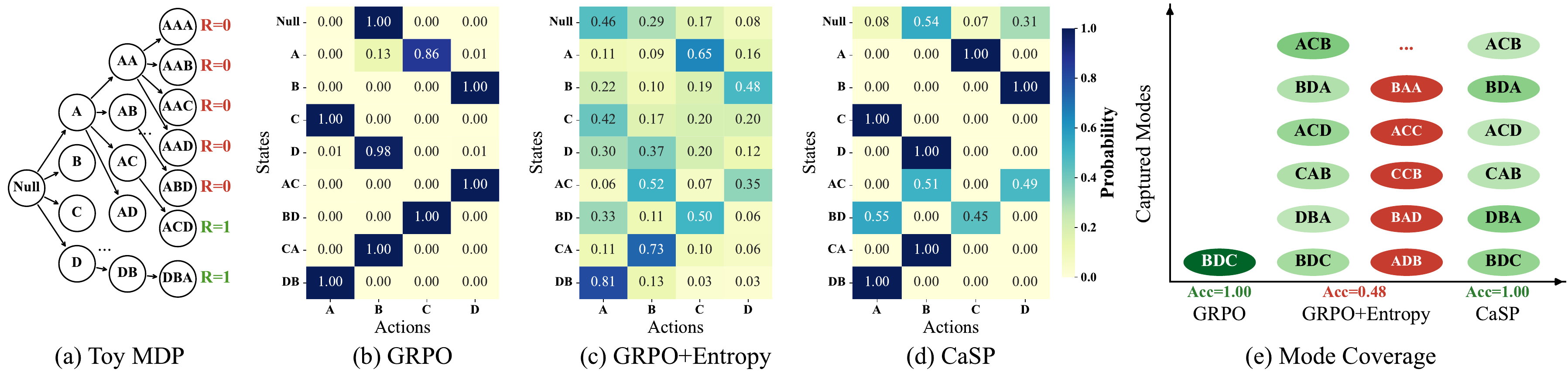}
\caption{\textbf{Toy example: mode collapse in GRPO, accuracy degradation in GRPO+Entropy, and diversity preservation in CaSP.}
  (a) Tree MDP with horizon 3 and action space \{A, B, C, D\}; rewarding sequences: ACD, ACB, BDC, BDA, CAB, DBA.
  (b) GRPO collapses to a single mode.
  (c) GRPO+Entropy recovers all modes but leaks mass to non-rewarding actions (red borders),
  accuracy = 0.48.
  (d) CaSP preserves all modes without leaking to dead-end actions, accuracy = 1.00.
  (e) Mode coverage and accuracy over 1,000 responses.}
  \label{fig:simko-toy}
\end{figure*}

\noindent\textbf{Setup.}
LLM generation under RLVR is naturally formulated as a finite-horizon, tree-structured MDP \cite{he2025random}. We study the learning dynamics in a tabular instance, where the policy samples actions from a categorical distribution at each state and a programmatic verifier provides reward at terminal states. This controlled setting aims to interpret how each method changes exploration. The full configuration is in the caption of Figure~\ref{fig:simko-toy}.

\noindent\textbf{Results.}
\textbf{GRPO} collapses to a single mode (BDC), placing nearly all root probability on action $B$. \textbf{GRPO+Entropy} recovers all six modes by flattening the distribution, but its entropy bonus indiscriminately spreads mass to non-rewarding actions at intermediate states (red borders in panel c), and accuracy degrades to $0.48$. \textbf{CaSP} preserves all six modes while keeping accuracy at $1.00$, since top-$N$ redistribution confines exploration to plausible continuations rather than the full action set.

\noindent\textbf{Discussion.}
The desideratum from Section \ref{section3} asks for non-negligible probability mass on the top-$N$ candidates, and CaSP addresses this at the gradient level. Positive updates are spread among the top-$N$ candidates rather than concentrated on the sampled token, and a stronger penalty on the rank-$1$ candidate prevents the squeeze effect from pushing mass back. Panel~(d) verifies this, with every rewarding root action retaining support. Vanilla entropy regularization cannot match this behavior, since an entropy bonus rewards any spread of mass regardless of which candidates receive it and therefore leaks probability onto dead-end actions (red borders in panel c). As shown in Figure~\ref{fig:methods}-(c), two distributions with identical entropy can differ sharply, so entropy alone cannot distinguish them. CaSP avoids this by acting on the identity of the top-$N$ candidates rather than on a scalar proxy.

\section{Experiments}
\label{sec:exp}

\subsection{Setup}

\noindent \textbf{Models and Datasets.}  
For math tasks, we experiment with a diverse set of models, including Qwen2.5-32B \cite{DBLP:journals/corr/abs-2412-15115}, Qwen2.5-Math-7B \cite{yang2024qwen2}, and Llama3.2-3B-Instruct \cite{dubey2024llama}. The Qwen models are trained on the MATH (Level 3-5) dataset \cite{hendrycks2021measuring}, while the Llama model is trained on a combined dataset of GSM8K \cite{cobbe2021training} and MATH (Level 1). For logical reasoning tasks, we train Qwen2.5-7B \cite{DBLP:journals/corr/abs-2412-15115} on the Synlogic-easy dataset \cite{liu2025synlogic}. For code reasoning, we train DeepSeek-R1-Distill-Qwen-7B on DeepCoder training dataset \cite{deepcoder2025}.

\noindent \textbf{Evaluation Protocol.}
We compare CaSP against several competitive baselines, including GRPO, PSR, NSR, with entropy loss, W-REINFORCE \cite{zhu2025surprising}, KL-Cov \cite{cui2025entropy}, with ``forking" tokens \citep{wang2025beyond}, P@k T. \cite{chen2025pass}, and Entropy-Adv \citep{cheng2025reasoning}. Evaluations are conducted on a variety of reasoning benchmarks: MATH500 \cite{hendrycks2021measuring}, Minerva\_math \cite{lewkowycz2022solving}, Olympiad-Bench \cite{he2024olympiadbench}, AMC, AIME, Synlogic-easy (validation split) \cite{liu2025synlogic}, BBH \cite{suzgun2022challenging},  LiveCodeBench \cite{DBLP:conf/iclr/JainHGLYZWSSS25}, and HumanEval+ \cite{DBLP:conf/nips/LiuXW023}. See more training and evaluation details in Appendix~\ref{appendix:training_detail} and \ref{appendix:evaluation_detail}.

\begin{table*}[!ht]
\caption{Pass@1/256 results for Qwen2.5-Math-7B and Llama3.2-3B-Instruct, and pass@1/128 results for Qwen2.5-32B on MATH500, AIME 2024/25, Minerva\_math, OlympiadBench, and AMC23.}
\label{math-result}
\centering
\setlength{\abovecaptionskip}{4.1pt}
\setlength{\belowcaptionskip}{-7pt}
\renewcommand{\arraystretch}{1.3}
\setlength{\tabcolsep}{4.7pt}    
\footnotesize
\begin{tabular}{lcccccc|c}
\textbf{Method} & \textbf{AIME24} & \textbf{AIME25} & \textbf{AMC23} & \textbf{MATH500}& \textbf{Minerva} & \textbf{Olympiad} & \textbf{Avg.} \\
\hline
\multicolumn{8}{l}{~~~~\textbf{\emph{Qwen2.5-Math-7B}}}\\
\hline
Base Model & 13.2 / 66.0 & 5.4 / 51.8 & 38.2 / 98.5 & 55.8 / 96.0 & 16.5 / 68.8 & 25.6 / 77.0 & 25.8 / 76.4 \\
GRPO & 28.1 / 72.3 & 11.5 / 52.1 & 61.2 / 97.1 & 76.6 / 96.2 & 33.4 / 64.0 & 39.1 / 74.7 & 41.7 / 76.1 \\
PSR & 19.3 / 68.5 & 11.2 / 48.9 & 62.1 / 94.9 & 74.0 / 91.4 & 32.8 / 63.6 & 37.6 / 67.7 & 39.5 / 72.5 \\
NSR & 22.8 / 80.3 & 9.7 / 61.2 & 59.4 / 100.0 & 74.6 / 97.0 & 32.9 / 65.1 & 37.8 / 78.4 & 39.5 / 80.3 \\
W-REINFORCE & 29.2 / 86.5 & 10.8 / 55.7 & 61.1 / 97.4 & 76.4 / 96.4 & 33.4 / 67.6 & 38.1 / 77.6 & 41.5 / 80.2 \\
KL-Cov & 30.9 / 81.2 & 11.7 / 55.2 & 62.2 / 97.4 & 76.5 / 97.0 & 34.4 / 66.2 & 39.2 / 76.9 & 42.5 / 79.0 \\
P@k T. & 26.7 / 77.9 & 10.2 / 61.3 & 58.8 / 97.5 & 73.3 / 96.8 &33.2 / 68.8 & 36.6 / 78.2 & 39.8 / 80.1 \\
GRPO w/ Entropy loss & 26.3 / 72.2 & 9.1 / 55.1 & 56.4 / 99.6 & 75.1 / 97.0 & 34.9 / 68.0 & 37.0 / 76.9 & 39.8 / 78.1 \\
GRPO w/ Entropy-Adv & 29.1 / 81.7 & 10.9 / 55.0 & 62.5 / 92.1 & 77.1 / 95.0 &33.5 / 60.7 & 39.7 / 71.9 & 42.1 / 76.1 \\
GRPO w/ forking tokens &28.6 / 74.6& 11.5 / 57.4& 59.6 / 96.7& 77.4 / 94.4& 33.9 / 65.1& 39.6 / 72.4& 41.8 / 76.8 \\
\rowcolor{Gray}CaSP & 32.8 / 78.0 & 12.9 / 64.6 & 62.4 / 97.5 & 77.6 / 96.8 & 35.0 / 68.4 & 39.8 / 77.8 & \textbf{43.4 / 80.5} \\
\rowcolor{Gray} $\Delta\text{(CaSP-GRPO)}$&\color{darkspringgreen} +4.7 / +5.7 &\color{darkspringgreen} +1.4 / +12.5 &\color{darkspringgreen} +1.2 / +0.4 &\color{darkspringgreen} +1.0 / +0.6 &\color{darkspringgreen} +1.6 / +4.4 &\color{darkspringgreen}\color{darkspringgreen} +0.7 / +3.1 &\color{darkspringgreen} +1.7 / +4.4 \\
\hline
\multicolumn{8}{l}{~~~~\textbf{\emph{Qwen2.5-32B}}}\\
\hline
Base Model & 9.7 / 63.3 & 4.9 / 50.0 & 42.1 / 100.0 & 64.5 / 97.0 & 27.0 / 67.6 & 28.1 / 77.3 & 29.4 / 75.9 \\
GRPO & 27.3 / 66.7 & 16.1 / 43.3 & 65.1 / 95.0 & 82.7 / 93.8 & 43.1 / 61.4 & 44.5 / 69.3 & 46.5 / 71.6 \\
GRPO w/ Entropy loss& 25.3 / 66.7 & 15.2 / 36.7 & 62.0 / 95.0 & 82.7 / 95.2 & 42.7 / 60.3 & 44.8 / 68.4 & 45.4 / 70.4 \\
\rowcolor{Gray}CaSP ($\alpha=0.01$) & 29.9 / 73.3 & 17.8 / 50.0 & 67.5 / 97.5 & 83.3 / 97.0 & 43.9 / 65.8 & 46.2 / 73.6 & 48.1 / 76.2 \\
\rowcolor{Gray}CaSP ($\alpha=0.05$) & 31.8 / 70.0 & 17.8 / 53.3 & 68.5 / 97.5 & 83.8 / 96.8 & 45.2 / 67.3 & 47.1 / 74.2 & \textbf{49.0 / 76.5} \\
\rowcolor{Gray} $\Delta\text{(CaSP-GRPO)}$&\color{darkspringgreen} +4.5 / \color{darkspringgreen}+3.3 &\color{darkspringgreen} +1.7 / +10 &\color{darkspringgreen} +3.4 / +2.5 &\color{darkspringgreen} +1.1 / \color{darkspringgreen}+3.0 & \color{darkspringgreen}+2.1 / \color{darkspringgreen}+5.9 &\color{darkspringgreen} +2.6 / \color{darkspringgreen}+4.9 &\color{darkspringgreen} +2.5 / +4.9 \\
\hline
\multicolumn{8}{l}{~~~~\textbf{\emph{Llama3.2-3B-Instruct}}}\\
\hline
Base Model & 3.4 / 51.7 & 0.7 / 46.7 & 20.3 / 94.9 & 37.8 / 93.6 & 10.1 / 59.2 & 12.7 / 67.1 & 14.2 / 68.9 \\
GRPO & 12.7 / 55.1 & 1.1 / 44.1 & 32.5 / 96.7 & 53.1 / 91.6 & 17.3 / 62.5 & 20.1 / 67.0 & 23.3 / 69.5 \\
PSR & 7.8 / 57.4 & 1.0 / 35.1 & 27.2 / 98.8 & 50.3 / 91.0 & 18.5 / 61.0 & 18.9 / 63.7 & 20.6 / 67.8 \\
NSR & 11.1 / 53.7 & 1.5 / 47.4 & 30.3 / 94.6 & 53.3 / 94.0 & 19.0 / 60.3 & 20.0 / 68.0 & 22.5 / 69.7 \\
W-REINFORCE & 13.3 / 51.7 & 1.1 / 42.1 & 31.4 / 96.3 & 52.4 / 92.8 & 16.7 / 59.9 & 19.6 / 65.8 & 22.4 / 68.1 \\
\rowcolor{Gray}CaSP & 13.8 / 54.6 & 1.0 / 45.4 & 35.2 / 98.8 & 54.6 / 93.4 & 18.5 / 63.2 & 21.0 / 69.6 & \textbf{24.0 / 70.8} \\
\rowcolor{Gray} $\Delta\text{(CaSP-GRPO)}$&\color{darkspringgreen} +1.1 / \color{darkspringred}-0.5 &\color{darkspringred}-0.1 / \color{darkspringgreen}+1.3 &\color{darkspringgreen} +2.7 / +2.1 &\color{darkspringgreen} +1.5 / +1.8 &\color{darkspringgreen}+1.2 / +0.7 &\color{darkspringgreen} +0.9 / +2.6 &\color{darkspringgreen} +0.7 / +1.3 \\
\end{tabular}
\end{table*}

\subsection{Math Reasoning}
\label{sec:math}

We evaluate CaSP on 6 math benchmarks across 3 backbones in Table~\ref{math-result}. Compared to the base models, CaSP significantly improves the pass@1 score by 17.6\% on Qwen2.5-Math-7B and 9.8\% on Llama3.2-3B-Instruct. At the same time, it also boosts the pass@256 score by 4.1\% and 1.9\% on the same backbones respectively, demonstrating improved exploration and overall reasoning quality. Notably, on the larger Qwen2.5-32B model, CaSP further improves pass@1 by 19.6\% and pass@128 by 0.6\%, suggesting that CaSP remains effective when scaled to larger backbones.

Compared to GRPO and other baselines (KL-Cov, Entropy-Adv, w/ ``forking'' tokens and P@k T.), CaSP consistently outperforms them across all backbones and $K$. More importantly, CaSP delivers these gains without sacrificing exploration (pass@256) and with even stronger exploitation (pass@1). Compared to GRPO, CaSP improves pass@256 by 4.4\% on Qwen2.5-Math-7B and 1.3\% on Llama3.2-3B-Instruct, and improves pass@128 by 4.9\% on Qwen2.5-32B, while also achieving higher pass@1 across all three backbones. Token-level variants provide less consistent gains: KL-Cov slightly improves pass@1 over GRPO (42.5\% vs. 41.7\%) but remains below CaSP in pass@256 (79.0\% vs. 80.5\%), while Entropy-Adv (42.1\% / 76.1\%) and training with ``forking'' tokens (41.8\% / 76.8\%) do not improve pass@256 over GRPO. Vanilla entropy loss encourages indiscriminate spreading rather than preserving plausible branches. This limitation is evident in its lower pass@$1$ compared with GRPO on Qwen2.5-Math-7B (39.8\% vs.\ 41.7\%) and Qwen2.5-32B (45.4\% vs.\ 46.5\%).

For NSR and W-REINFORCE, strong pass@256 performance is maintained, but often at the expense of much lower pass@1. In contrast, CaSP achieves a better balance on most backbones. On Qwen2.5-Math-7B, CaSP reaches a slightly higher pass@256 score (80.5\% vs. 80.3\% for NSR and 80.2\% for W-REINFORCE) while clearly outperforming both in pass@1 (43.4\% vs. 39.5\% and 41.5\%). A similar trend is observed on Llama3.2-3B-Instruct, where CaSP improves both pass@256 (70.8\% vs. 69.7\% and 68.1\%) and pass@1 (24.0\% vs. 22.5\% and 22.4\%). 
These results support our hypothesis that alleviating probability over-concentration improves pass@$K$ performance, indicating a better balance between exploitation and exploration.

\begin{table*}[t]
\centering
\setlength{\abovecaptionskip}{4.1pt}
\setlength{\belowcaptionskip}{-7pt}
\caption{Pass@$K$ results for Qwen2.5-7B on Synlogic and BBH Datasets.}
\label{logic-result}
\setlength{\tabcolsep}{4pt}   
\footnotesize

\renewcommand{\arraystretch}{1.19}
\begin{tabular}{c|cccccccc|cccccccc}
\multirow{2}{*}{\textbf{Method}} & \multicolumn{8}{c|}
{\textbf{Synlogic}} & \multicolumn{8}{c}{\textbf{BBH}} \\ 
 & 1    & 2    & 4    & 8    & 16   & 32   & 64   & 128  & 1    & 2    & 4    & 8    & 16   & 32   & 64   & 128  \\ 
\hline 
  Base Model  & 3.1 & 4.9 & 7.5 & 11.1 & 15.4 & 20.0 & 24.5 & 28.7 & 42.4 & 59.3 & \underline{74.4} & \textbf{84.6} & \textbf{89.9} & \textbf{92.4} & \textbf{93.6} & \textbf{94.2}    \\ 
  GRPO   & \underline{34.0} & \underline{36.7} & \underline{39.1}& \underline{41.1} & \underline{43.1} & \underline{45.0} & \underline{47.0} & \underline{48.6} &  54.4 & \underline{62.8} & 69.8 & 75.6& 79.3 & 82.5 & 84.9 & 86.8  \\ 
PSR  & 27.3 & 29.3 & 31.7 & 34.3 & 36.9 & 39.4 & 41.6 & 43.6 & \underline{54.9} & 62.7 & 68.8 & 73.4 & 76.8 & 79.4 & 81.4 & 82.8    \\ 
W-REINFORCE & 0.8 & 1.3& 1.8 &2.4& 2.9&3.4&3.8&3.9 & 15.4 & 21.9 & 27.8 & 32.3 & 35.4 & 37.2 & 38.3 & 38.8 \\\rowcolor{Gray}
CaSP & \textbf{34.6} & \textbf{38.3} & \textbf{41.8} & \textbf{45.3} & \textbf{48.1} & \textbf{50.7} & \textbf{53.1} & \textbf{55.1} &  \textbf{58.4} & \textbf{69.2} & \textbf{77.3} & \underline{83.0} & \underline{86.8} & \underline{89.3} & \underline{91.0} & \underline{92.1} \\
\end{tabular}
\vspace{-3mm}
\end{table*}

\begin{table}[!ht]
\centering

\caption{ Performance comparison on code reasoning benchmarks for DeepSeek-R1-Distill-Qwen-7B.}
\label{code-result}

\renewcommand{\arraystretch}{1.19}
\setlength{\tabcolsep}{5pt}      
\footnotesize
\begin{tabular}{l|cccc}
\textbf{Method} 
& \multicolumn{2}{c}{\textbf{LiveCodeBench}} 
& \multicolumn{2}{c}{\textbf{HumanEval+}} \\
& Avg@16 & Pass@16 
& Avg@16 & Pass@16 \\
\midrule
Base Model & 30.85 & 47.31  & 81.02 & 94.48 \\
GRPO       & 33.92 & 53.41  & 80.25 & 94.48 \\
\rowcolor{Gray} CaSP & \textbf{36.02} & \textbf{54.48}  & \textbf{83.90} & \textbf{96.32} \\
\end{tabular}
\vspace{-2mm}
\end{table}

\subsection{Logical and Coding Reasoning}
\label{sec:logical_code}

We evaluate the generalization of CaSP on two logic reasoning and two code reasoning benchmarks, as shown in Table~\ref{code-result} and Table~\ref{logic-result}. 

On Synlogic, CaSP significantly outperforms the base model, with a +31.5\% gain in pass@1 and +26.4\% at pass@128. GRPO and PSR show improvements but lag behind CaSP by 6.5\% and 11.5\% at pass@128. W-REINFORCE, however, fails to train effectively, with pass@1 scores of only 0.8\%. Similar observations hold on BBH.

On BBH, CaSP boosts the base model's pass@1 to 58.4\% (+16.0\%), and maintains stability at higher sampling rates, with a small 2.1\% decrease in pass@128. GRPO and PSR, by comparison, drop 7.4\% and 11.4\% at pass@128 compared to the base model, showing difficulties in sustaining performance. W-REINFORCE performs poorly, achieving only 15.4\% at pass@1. These results demonstrate that relying solely on negative samples is insufficient to improve pass@$K$ on challenging tasks. In contrast, CaSP exhibits strong generalization, effectively trains on difficult tasks, and improves pass@$K$ performance by mitigating probability over-concentration.

Table~\ref{code-result} presents the results on code reasoning benchmarks. On LiveCodeBench, CaSP achieves significant improvements, gaining +5.17\% in pass@1 and +7.17\% in pass@16, and outperforming GRPO by +2.1\% and +1.07\%, respectively. On HumanEval+, CaSP achieves the best performance with an Avg@16 score of 83.90\%, while GRPO even degrades compared to the base model.

\begin{figure}[t]
    \centering
    \setlength{\abovecaptionskip}{2pt}
\setlength{\belowcaptionskip}{-6pt}
    \includegraphics[width=1\linewidth]{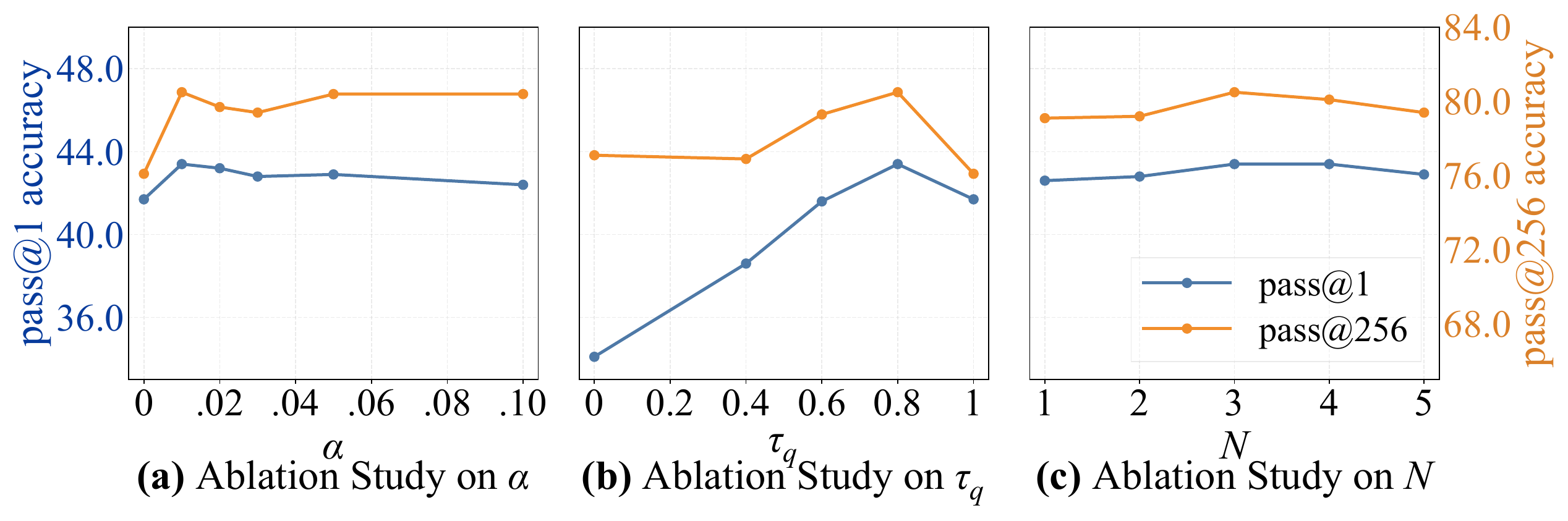}
    \caption{Ablations on $\alpha$, $\tau_q$, and $N$ using Qwen2.5-Math-7B on math benchmarks.}
    \label{fig:ablation}
    \vspace{-2mm}
\end{figure}

\subsection{Ablations}

We ablate the hyperparameters $\alpha$, $\tau$, and $N$, as well as the contributions of the positive and negative components. The full ablation results are  in Appendix~\ref{appendix:ablation result}, and the performance variations with respect to $\alpha$, $\tau$, and $N$ are shown in Figure~\ref{fig:ablation}.

We evaluate $\alpha$ values from 0 to 0.1, with $\alpha = 0$ representing the performance of GRPO. Increasing $\alpha$ results in a monotonic improvement in pass@256, with gains ranging from 3.3\% to 4.4\% compared to GRPO. In contrast, pass@1 performance peaks at $\alpha = 0.01$ and then slightly degrades, though it remains superior to GRPO.

For $\tau$, we sweep from $\tau_0$ (all tokens receive CaSP updates) to $\tau_1$ (no tokens, equivalent to GRPO). Notably, CaSP outperforms GRPO in pass@256 across most of $\tau_q$. However, when CaSP is applied to all tokens, pass@1 drops by 9.3\%. This indicates that CaSP should target high-entropy tokens instead of uniformly smoothing all.

For $N$, we test values from 1 to 5. As $N$ increases, both pass@256 and pass@1 show an initial increase followed by a decrease. This trend suggests that restricting optimization to a small subset of the most probable candidates is sufficient. Specifically, pass@256 stays between 79.1\% and 80.5\%, while pass@1 fluctuates between 42.6\% and 43.4\%, both metrics outperforming GRPO.

\section{Conclusion}
We revisited exploration collapse in RLVR through the candidate-level distribution. Both positive and negative updates concentrate mass on the rank-$1$ candidate, collapsing rollout diversity regardless of $K$. To preserve mass on the top-$N$ candidates, we proposed CaSP, which redistributes positive gradients across the top-$N$ and applies a stronger penalty to the rank-$1$ candidate. CaSP improves pass@$1$ and pass@$K$ simultaneously over GRPO and other strong baselines across math, logical, and coding benchmarks, with gains persisting up to 32B parameters and $K=1024$.

\clearpage
\section{Limitations}
CaSP improves exploration by redistributing gradients within the local top-$N$ candidate set, but this design also makes it dependent on the quality of the model's current candidate ranking. If useful reasoning branches are not assigned sufficient probability to enter the top-$N$ set, CaSP cannot directly recover them. This limitation arises from a deliberate efficiency trade-off: operating on the full vocabulary would provide broader coverage but is computationally expensive and risks assigning probability mass to implausible tokens, while top-$N$ redistribution keeps the update tractable and targeted. Extending CaSP to adaptively expand or refresh the candidate set is an interesting direction for future work.

\section*{Ethical Considerations}
The study does not involve human subjects or the use of personal or sensitive data. All datasets and code utilized and released conform to their respective licenses and terms of use.  The contributions in this work are foundational and do not raise issues related to fairness, privacy, security, or potential misuse. We confirm that all ethical considerations have been thoroughly addressed.


\bibliography{latex/custom}

@inproceedings{DBLP:conf/nips/LiuXW023,
  author       = {Jiawei Liu and
                  Chunqiu Steven Xia and
                  Yuyao Wang and
                  Lingming Zhang},
  editor       = {Alice Oh and
                  Tristan Naumann and
                  Amir Globerson and
                  Kate Saenko and
                  Moritz Hardt and
                  Sergey Levine},
  title        = {Is Your Code Generated by ChatGPT Really Correct? Rigorous Evaluation
                  of Large Language Models for Code Generation},
  booktitle    = {Advances in Neural Information Processing Systems 36: Annual Conference
                  on Neural Information Processing Systems 2023, NeurIPS 2023, New Orleans,
                  LA, USA, December 10 - 16, 2023},
  year         = {2023},
  url          = {http://papers.nips.cc/paper\_files/paper/2023/hash/43e9d647ccd3e4b7b5baab53f0368686-Abstract-Conference.html},
  bibsource    = {dblp computer science bibliography, https://dblp.org}
}

@inproceedings{DBLP:conf/iclr/JainHGLYZWSSS25,
  author       = {Naman Jain and
                  King Han and
                  Alex Gu and
                  Wen{-}Ding Li and
                  Fanjia Yan and
                  Tianjun Zhang and
                  Sida Wang and
                  Armando Solar{-}Lezama and
                  Koushik Sen and
                  Ion Stoica},
  title        = {LiveCodeBench: Holistic and Contamination Free Evaluation of Large
                  Language Models for Code},
  booktitle    = {The Thirteenth International Conference on Learning Representations,
                  {ICLR} 2025, Singapore, April 24-28, 2025},
  publisher    = {OpenReview.net},
  year         = {2025},
  url          = {https://openreview.net/forum?id=chfJJYC3iL},
  bibsource    = {dblp computer science bibliography, https://dblp.org}
}

@article{jiang2025rethinking,
  title={Rethinking Entropy Regularization in Large Reasoning Models},
  author={Jiang, Yuxian and Li, Yafu and Chen, Guanxu and Liu, Dongrui and Cheng, Yu and Shao, Jing},
  journal={arXiv preprint arXiv:2509.25133},
  year={2025}
}

@article{tachet2018learning,
  title={On the learning dynamics of deep neural networks},
  author={Tachet, Remi and Pezeshki, Mohammad and Shabanian, Samira and Courville, Aaron and Bengio, Yoshua},
  journal={arXiv preprint arXiv:1809.06848},
  year={2018}
}

@article{deng2025effect,
  title={On the Effect of Negative Gradient in Group Relative Deep Reinforcement Optimization},
  author={Deng, Wenlong and Ren, Yi and Li, Muchen and Sutherland, Danica J and Li, Xiaoxiao and Thrampoulidis, Christos},
  journal={arXiv preprint arXiv:2505.18830},
  year={2025}
}

@article{kang2024learning,
  title={What Do Learning Dynamics Reveal About Generalization in LLM Reasoning?},
  author={Kang, Katie and Setlur, Amrith and Ghosh, Dibya and Steinhardt, Jacob and Tomlin, Claire and Levine, Sergey and Kumar, Aviral},
  journal={arXiv preprint arXiv:2411.07681},
  year={2024}
}

@article{wang2025beyond,
  title={Beyond the 80/20 rule: High-entropy minority tokens drive effective reinforcement learning for llm reasoning},
  author={Wang, Shenzhi and Yu, Le and Gao, Chang and Zheng, Chujie and Liu, Shixuan and Lu, Rui and Dang, Kai and Chen, Xionghui and Yang, Jianxin and Zhang, Zhenru and others},
  journal={arXiv preprint arXiv:2506.01939},
  year={2025}
}

@article{walder2025pass,
  title={Pass@ k policy optimization: Solving harder reinforcement learning problems},
  author={Walder, Christian and Karkhanis, Deep Tejas},
  journal={Advances in Neural Information Processing Systems},
  volume={38},
  pages={152416--152445},
  year={2026}
}

@article{tajwar2026maximum,
  title={Maximum Likelihood Reinforcement Learning},
  author={Tajwar, Fahim and Zeng, Guanning and Zhou, Yueer and Song, Yuda and Arora, Daman and Jiang, Yiding and Schneider, Jeff and Salakhutdinov, Ruslan and Feng, Haiwen and Zanette, Andrea},
  journal={arXiv preprint arXiv:2602.02710},
  year={2026}
}

@article{li2026setpo,
  title={SetPO: Set-Level Policy Optimization for Diversity-Preserving LLM Reasoning},
  author={Li, Chenyi and Zhang, Yuan and Wang, Bo and Ma, Guoqing and Tang, Wei and Huang, Haoyang and Duan, Nan},
  journal={arXiv preprint arXiv:2602.01062},
  year={2026}
}

@article{li2025jointly,
  title={Jointly reinforcing diversity and quality in language model generations},
  author={Li, Tianjian and Zhang, Yiming and Yu, Ping and Saha, Swarnadeep and Khashabi, Daniel and Weston, Jason and Lanchantin, Jack and Wang, Tianlu},
  journal={arXiv preprint arXiv:2509.02534},
  year={2025}
}

@article{wandsdr,
  author       = {Zhongwei Wan and
                  Yun Shen and
                  Zhihao Dou and
                  Donghao Zhou and
                  Yu Zhang and
                  Xin Wang and
                  Hui Shen and
                  Jing Xiong and
                  Chaofan Tao and
                  Zixuan Zhong and
                  Peizhou Huang and
                  Mi Zhang},
  title        = {{DSDR:} Dual-Scale Diversity Regularization for Exploration in {LLM}
                  Reasoning},
  journal      = {CoRR},
  volume       = {abs/2602.19895},
  year         = {2026},
  url          = {https://doi.org/10.48550/arXiv.2602.19895},
  doi          = {10.48550/ARXIV.2602.19895},
  eprinttype   = {arXiv},
  eprint       = {2602.19895},
  bibsource    = {dblp computer science bibliography, https://dblp.org}
}

@article{yangdcpo,
  author       = {Shihui Yang and
                  Chengfeng Dou and
                  Peidong Guo and
                  Kai Lu and
                  Qiang Ju and
                  Fei Deng and
                  Rihui Xin},
  title        = {{DCPO:} Dynamic Clipping Policy Optimization},
  journal      = {CoRR},
  volume       = {abs/2509.02333},
  year         = {2025},
  url          = {https://doi.org/10.48550/arXiv.2509.02333},
  doi          = {10.48550/ARXIV.2509.02333},
  eprinttype   = {arXiv},
  eprint       = {2509.02333},
  bibsource    = {dblp computer science bibliography, https://dblp.org}
}

@article{lin2026resrl,
  title={Resrl: Boosting llm reasoning via negative sample projection residual reinforcement learning},
  author={Lin, Zihan and Wang, Xiaohan and Cao, Jie and Chai, Jiajun and Wang, Li and Lu, Xiaodong and Lin, Wei and He, Ran and Yin, Guojun},
  journal={arXiv preprint arXiv:2605.00380},
  year={2026}
}

@article{zhang2025rspo,
  title={Rspo: Risk-seeking policy optimization for pass@ k and max@ k metrics in large language models},
  author={Zhang, Kaichen and Gao, Shenghao and Hong, Yuzhong and Sun, Haipeng and Bao, Junwei and Jiang, Hongfei and Song, Yang and Dingqian, Hong and Xiong, Hui},
  journal={arXiv preprint arXiv:2508.01174},
  year={2025}
}

@article{chen2025pass,
  title={Pass@ k training for adaptively balancing exploration and exploitation of large reasoning models},
  author={Chen, Zhipeng and Qin, Xiaobo and Wu, Youbin and Ling, Yue and Ye, Qinghao and Zhao, Wayne Xin and Shi, Guang},
  journal={arXiv preprint arXiv:2508.10751},
  year={2025}
}

@article{yang2025not,
  title={Do Not Let Low-Probability Tokens Over-Dominate in RL for LLMs},
  author={Yang, Zhihe and Luo, Xufang and Wang, Zilong and Han, Dongqi and He, Zhiyuan and Li, Dongsheng and Xu, Yunjian},
  journal={arXiv preprint arXiv:2505.12929},
  year={2025}
}

@article{cobbe2021training,
  title={Training verifiers to solve math word problems},
  author={Cobbe, Karl and Kosaraju, Vineet and Bavarian, Mohammad and Chen, Mark and Jun, Heewoo and Kaiser, Lukasz and Plappert, Matthias and Tworek, Jerry and Hilton, Jacob and Nakano, Reiichiro and others},
  journal={arXiv preprint arXiv:2110.14168},
  year={2021}
}

@article{yang2024qwen2,
  title={Qwen2. 5-math technical report: Toward mathematical expert model via self-improvement},
  author={Yang, An and Zhang, Beichen and Hui, Binyuan and Gao, Bofei and Yu, Bowen and Li, Chengpeng and Liu, Dayiheng and Tu, Jianhong and Zhou, Jingren and Lin, Junyang and others},
  journal={arXiv preprint arXiv:2409.12122},
  year={2024}
}

@article{DBLP:journals/corr/abs-2412-15115,
  author       = {An Yang and Baosong Yang and Beichen Zhang and Binyuan Hui and Bo Zheng and Bowen Yu and Chengyuan Li and Dayiheng Liu and Fei Huang and Haoran Wei and Huan Lin and Jian Yang and Jianhong Tu and Jianwei Zhang and Jianxin Yang and Jiaxi Yang and Jingren Zhou and Junyang Lin and Kai Dang and Keming Lu and Keqin Bao and Kexin Yang and Le Yu and Mei Li and Mingfeng Xue and Pei Zhang and Qin Zhu and Rui Men and Runji Lin and Tianhao Li and Tianyi Tang and Tingyu Xia and Xingzhang Ren and Xuancheng Ren and Yang Fan and Yang Su and Yichang Zhang and Yu Wan and Yuqiong Liu and Zeyu Cui and Zhenru Zhang and Zihan Qiu},
  title        = {Qwen2.5 Technical Report},
  journal={arXiv preprint arXiv:2412.15115},
  year={2024}
}

@article{dubey2024llama,
  title={The llama 3 herd of models},
  author={Dubey, Abhimanyu and Jauhri, Abhinav and Pandey, Abhinav and Kadian, Abhishek and Al-Dahle, Ahmad and Letman, Aiesha and Mathur, Akhil and Schelten, Alan and Yang, Amy and Fan, Angela and others},
  journal={arXiv e-prints},
  pages={arXiv--2407},
  year={2024}
}

@article{liu2025synlogic,
  title={SynLogic: Synthesizing Verifiable Reasoning Data at Scale for Learning Logical Reasoning and Beyond},
  author={Liu, Junteng and Fan, Yuanxiang and Jiang, Zhuo and Ding, Han and Hu, Yongyi and Zhang, Chi and Shi, Yiqi and Weng, Shitong and Chen, Aili and Chen, Shiqi and others},
  journal={arXiv preprint arXiv:2505.19641},
  year={2025}
}

@article{suzgun2022challenging,
  title={Challenging big-bench tasks and whether chain-of-thought can solve them},
  author={Suzgun, Mirac and Scales, Nathan and Sch{\"a}rli, Nathanael and Gehrmann, Sebastian and Tay, Yi and Chung, Hyung Won and Chowdhery, Aakanksha and Le, Quoc V and Chi, Ed H and Zhou, Denny and others},
  journal={arXiv preprint arXiv:2210.09261},
  year={2022}
}

@article{li2025questa,
  title={Questa: Expanding reasoning capacity in llms via question augmentation},
  author={Li, Jiazheng and Lu, Hong and Wen, Kaiyue and Yang, Zaiwen and Gao, Jiaxuan and Lin, Hongzhou and Wu, Yi and Zhang, Jingzhao},
  journal={arXiv preprint arXiv:2507.13266},
  year={2025}
}

@article{he2025skywork,
  title={Skywork open reasoner 1 technical report},
  author={He, Jujie and Liu, Jiacai and Liu, Chris Yuhao and Yan, Rui and Wang, Chaojie and Cheng, Peng and Zhang, Xiaoyu and Zhang, Fuxiang and Xu, Jiacheng and Shen, Wei and others},
  journal={arXiv preprint arXiv:2505.22312},
  year={2025}
}

@article{hou2025advancing,
  title={Advancing language model reasoning through reinforcement learning and inference scaling},
  author={Hou, Zhenyu and Lv, Xin and Lu, Rui and Zhang, Jiajie and Li, Yujiang and Yao, Zijun and Li, Juanzi and Tang, Jie and Dong, Yuxiao},
  journal={arXiv preprint arXiv:2501.11651},
  year={2025}
}

@article{wu2025invisible,
  title={The invisible leash: Why rlvr may not escape its origin},
  author={Wu, Fang and Xuan, Weihao and Lu, Ximing and Harchaoui, Zaid and Choi, Yejin},
  journal={arXiv preprint arXiv:2507.14843},
  year={2025}
}

@article{zheng2025group,
  title={Group sequence policy optimization},
  author={Zheng, Chujie and Liu, Shixuan and Li, Mingze and Chen, Xiong-Hui and Yu, Bowen and Gao, Chang and Dang, Kai and Liu, Yuqiong and Men, Rui and Yang, An and others},
  journal={arXiv preprint arXiv:2507.18071},
  year={2025}
}

@article{dong2025rl,
  title={Rl-plus: Countering capability boundary collapse of llms in reinforcement learning with hybrid-policy optimization},
  author={Dong, Yihong and Jiang, Xue and Tao, Yongding and Liu, Huanyu and Zhang, Kechi and Mou, Lili and Cao, Rongyu and Ma, Yingwei and Chen, Jue and Li, Binhua and others},
  journal={arXiv preprint arXiv:2508.00222},
  year={2025}
}

@article{liang2025beyond,
  title={Beyond pass@ 1: Self-play with variational problem synthesis sustains rlvr},
  author={Liang, Xiao and Li, Zhongzhi and Gong, Yeyun and Shen, Yelong and Wu, Ying Nian and Guo, Zhijiang and Chen, Weizhu},
  journal={arXiv preprint arXiv:2508.14029},
  year={2025}
}

@article{cheng2025reasoning,
  title={Reasoning with exploration: An entropy perspective},
  author={Cheng, Daixuan and Huang, Shaohan and Zhu, Xuekai and Dai, Bo and Zhao, Wayne Xin and Zhang, Zhenliang and Wei, Furu},
  journal={arXiv preprint arXiv:2506.14758},
  year={2025}
}

@inproceedings{
    ren2024learning,
    title={Learning Dynamics of {LLM} Finetuning},
    author={Yi Ren and Danica J. Sutherland},
    booktitle={The Thirteenth International Conference on Learning Representations},
    year={2025},
    url={https://openreview.net/forum?id=tPNHOoZFl9}
}

@article{lewkowycz2022solving,
  title={Solving quantitative reasoning problems with language models},
  author={Lewkowycz, Aitor and Andreassen, Anders and Dohan, David and Dyer, Ethan and Michalewski, Henryk and Ramasesh, Vinay and Slone, Ambrose and Anil, Cem and Schlag, Imanol and Gutman-Solo, Theo and others},
  journal={Advances in neural information processing systems},
  volume={35},
  pages={3843--3857},
  year={2022}
}

@article{he2024olympiadbench,
  title={Olympiadbench: A challenging benchmark for promoting agi with olympiad-level bilingual multimodal scientific problems},
  author={He, Chaoqun and Luo, Renjie and Bai, Yuzhuo and Hu, Shengding and Thai, Zhen Leng and Shen, Junhao and Hu, Jinyi and Han, Xu and Huang, Yujie and Zhang, Yuxiang and others},
  journal={arXiv preprint arXiv:2402.14008},
  year={2024}
}

@article{hendrycks2021measuring,
  title={Measuring mathematical problem solving with the math dataset},
  author={Hendrycks, Dan and Burns, Collin and Kadavath, Saurav and Arora, Akul and Basart, Steven and Tang, Eric and Song, Dawn and Steinhardt, Jacob},
  journal={arXiv preprint arXiv:2103.03874},
  year={2021}
}

@inproceedings{zeng2025simplerlzooinvestigatingtamingzero,
      title={SimpleRL-Zoo: Investigating and Taming Zero Reinforcement Learning for Open Base Models in the Wild}, 
      author={Weihao Zeng and Yuzhen Huang and Qian Liu and Wei Liu and Keqing He and Zejun Ma and Junxian He},
      booktitle={Second Conference on Language Modeling},
      year={2025}
}

@misc{deepseekai2025deepseekr1incentivizingreasoningcapability,
      title={DeepSeek-R1: Incentivizing Reasoning Capability in LLMs via Reinforcement Learning}, 
      author={DeepSeek-AI},
      year={2025},
      eprint={2501.12948},
      archivePrefix={arXiv},
      primaryClass={cs.CL},
      url={https://arxiv.org/abs/2501.12948}, 
}

@article{shao2024deepseekmath,
  title={Deepseekmath: Pushing the limits of mathematical reasoning in open language models},
  author={Shao, Zhihong and Wang, Peiyi and Zhu, Qihao and Xu, Runxin and Song, Junxiao and Bi, Xiao and Zhang, Haowei and Zhang, Mingchuan and Li, YK and Wu, Yang and others},
  journal={arXiv preprint arXiv:2402.03300},
  year={2024}
}

@article{zhu2025surprising,
  title={The surprising effectiveness of negative reinforcement in LLM reasoning},
  author={Zhu, Xinyu and Xia, Mengzhou and Wei, Zhepei and Chen, Wei-Lin and Chen, Danqi and Meng, Yu},
  journal={arXiv preprint arXiv:2506.01347},
  year={2025}
}

@article{muller2019does,
  title={When does label smoothing help?},
  author={M{\"u}ller, Rafael and Kornblith, Simon and Hinton, Geoffrey E},
  journal={Advances in neural information processing systems},
  volume={32},
  year={2019}
}

@article{schulman2017proximal,
  title={Proximal policy optimization algorithms},
  author={Schulman, John and Wolski, Filip and Dhariwal, Prafulla and Radford, Alec and Klimov, Oleg},
  journal={arXiv preprint arXiv:1707.06347},
  year={2017}
}

@article{yang2025depth,
  title={Depth-breadth synergy in rlvr: Unlocking llm reasoning gains with adaptive exploration},
  author={Yang, Zhicheng and Guo, Zhijiang and Huang, Yinya and Wang, Yongxin and Xie, Dongchun and Wang, Yiwei and Liang, Xiaodan and Tang, Jing},
  journal={arXiv preprint arXiv:2508.13755},
  year={2025}
}

@article{yu2025dapo,
  title={Dapo: An open-source llm reinforcement learning system at scale},
  author={Yu, Qiying and Zhang, Zheng and Zhu, Ruofei and Yuan, Yufeng and Zuo, Xiaochen and Yue, Yu and Dai, Weinan and Fan, Tiantian and Liu, Gaohong and Liu, Lingjun and others},
  journal={arXiv preprint arXiv:2503.14476},
  year={2025}
}

@article{liu2025understanding,
  title={Understanding r1-zero-like training: A critical perspective},
  author={Liu, Zichen and Chen, Changyu and Li, Wenjun and Qi, Penghui and Pang, Tianyu and Du, Chao and Lee, Wee Sun and Lin, Min},
  journal={arXiv preprint arXiv:2503.20783},
  year={2025}
}

@misc{hu2025openreasonerzeroopensourceapproach,
      title={Open-Reasoner-Zero: An Open Source Approach to Scaling Up Reinforcement Learning on the Base Model}, 
      author={Jingcheng Hu and Yinmin Zhang and Qi Han and Daxin Jiang and Xiangyu Zhang and Heung-Yeung Shum},
      year={2025},
      eprint={2503.24290},
      archivePrefix={arXiv},
      primaryClass={cs.LG},
      url={https://arxiv.org/abs/2503.24290}, 
}

@article{cui2025entropy,
  title={The entropy mechanism of reinforcement learning for reasoning language models},
  author={Cui, Ganqu and Zhang, Yuchen and Chen, Jiacheng and Yuan, Lifan and Wang, Zhi and Zuo, Yuxin and Li, Haozhan and Fan, Yuchen and Chen, Huayu and Chen, Weize and others},
  journal={arXiv preprint arXiv:2505.22617},
  year={2025}
}

@article{chu2025sft,
  title={Sft memorizes, rl generalizes: A comparative study of foundation model post-training},
  author={Chu, Tianzhe and Zhai, Yuexiang and Yang, Jihan and Tong, Shengbang and Xie, Saining and Schuurmans, Dale and Le, Quoc V and Levine, Sergey and Ma, Yi},
  journal={arXiv preprint arXiv:2501.17161},
  year={2025}
}

@article{hu2025reinforce++,
  title={Reinforce++: A simple and efficient approach for aligning large language models},
  author={Hu, Jian},
  journal={arXiv preprint arXiv:2501.03262},
  year={2025}
}

@article{he2025rewarding,
  title={Rewarding the Unlikely: Lifting GRPO Beyond Distribution Sharpening},
  author={He, Andre and Fried, Daniel and Welleck, Sean},
  journal={arXiv preprint arXiv:2506.02355},
  year={2025}
}

@article{song2025outcome,
  title={Outcome-based exploration for llm reasoning},
  author={Song, Yuda and Kempe, Julia and Munos, Remi},
  journal={arXiv preprint arXiv:2509.06941},
  year={2025}
}

@article{pal2024smaug,
  title={Smaug: Fixing failure modes of preference optimisation with dpo-positive},
  author={Pal, Arka and Karkhanis, Deep and Dooley, Samuel and Roberts, Manley and Naidu, Siddartha and White, Colin},
  journal={arXiv preprint arXiv:2402.13228},
  year={2024}
}

@article{yue2025does,
  title={Does reinforcement learning really incentivize reasoning capacity in llms beyond the base model?},
  author={Yue, Yang and Chen, Zhiqi and Lu, Rui and Zhao, Andrew and Wang, Zhaokai and Song, Shiji and Huang, Gao},
  journal={arXiv preprint arXiv:2504.13837},
  year={2025}
}

@article{deng2025token,
  title={Token Hidden Reward: Steering Exploration-Exploitation in Group Relative Deep Reinforcement Learning},
  author={Deng, Wenlong and Ren, Yi and Li, Yushu and Gong, Boying and Sutherland, Danica J and Li, Xiaoxiao and Thrampoulidis, Christos},
  journal={arXiv preprint arXiv:2510.03669},
  year={2025}
}

@article{ren2025thesis,
  title={Learning Dynamics of Deep Learning--Force Analysis of Deep Neural Networks},
  author={Ren, Yi},
  journal={arXiv preprint arXiv:2509.19554},
  year={2025}
}

@article{advani2020high,
  title={High-dimensional dynamics of generalization error in neural networks},
  author={Advani, Madhu S and Saxe, Andrew M and Sompolinsky, Haim},
  journal={Neural Networks},
  volume={132},
  pages={428--446},
  year={2020},
  publisher={Elsevier}
}

@article{saxe2013exact,
  title={Exact solutions to the nonlinear dynamics of learning in deep linear neural networks},
  author={Saxe, Andrew M and McClelland, James L and Ganguli, Surya},
  journal={arXiv preprint arXiv:1312.6120},
  year={2013}
}

@inproceedings{
razin2025unintentional,
title={Unintentional Unalignment: Likelihood Displacement in Direct Preference Optimization},
author={Noam Razin and Sadhika Malladi and Adithya Bhaskar and Danqi Chen and Sanjeev Arora and Boris Hanin},
booktitle={The Thirteenth International Conference on Learning Representations},
year={2025},
}

@article{dai2025cde,
  title={Cde: Curiosity-driven exploration for efficient reinforcement learning in large language models},
  author={Dai, Runpeng and Song, Linfeng and Liu, Haolin and Liang, Zhenwen and Yu, Dian and Mi, Haitao and Tu, Zhaopeng and Liu, Rui and Zheng, Tong and Zhu, Hongtu and others},
  journal={arXiv preprint arXiv:2509.09675},
  year={2025}
}

@article{gai2025differential,
  title={Differential smoothing mitigates sharpening and improves llm reasoning},
  author={Gai, Jingchu and Zeng, Guanning and Zhang, Huaqing and Raghunathan, Aditi},
  journal={arXiv preprint arXiv:2511.19942},
  year={2025}
}

@article{liu2026past,
  title={The Past Is Not Past: Memory-Enhanced Dynamic Reward Shaping},
  author={Liu, Yang and Wang, Enxi and Gao, Yufei and Zhang, Weixin and Wang, Bo and Zeng, Zhiyuan and Zhang, Yikai and Zheng, Yining and Qiu, Xipeng},
  journal={arXiv preprint arXiv:2604.11297},
  year={2026}
}

@misc{deepcoder2025,
  title={DeepCoder: A Fully Open-Source 14B Coder at O3-mini Level},
  author={Michael Luo and Sijun Tan and Roy Huang and Ameen Patel and Alpay Ariyak and Qingyang Wu and Xiaoxiang Shi and Rachel Xin and Colin Cai and Maurice Weber and Ce Zhang and Li Erran Li and Raluca Ada Popa and Ion Stoica},
  howpublished={\url{https://pretty-radio-b75.notion.site/DeepCoder-A-Fully-Open-Source-14B-Coder-at-O3-mini-Level-1cf81902c14680b3bee5eb349a512a51}},
  note={Notion Blog},
  year={2025}
}

@article{he2025random,
  title={Random policy valuation is enough for llm reasoning with verifiable rewards},
  author={He, Haoran and Ye, Yuxiao and Cai, Qingpeng and Hu, Chen and Jiao, Binxing and Jiang, Daxin and Pan, Ling},
  journal={arXiv preprint arXiv:2509.24981},
  year={2025}
}

@article{chen2025beyond,
  title={Beyond high-entropy exploration: Correctness-aware low-entropy segment-based advantage shaping for reasoning llms},
  author={Chen, Xinzhu and Li, Xuesheng and Sun, Zhongxiang and Yu, Weijie},
  journal={arXiv preprint arXiv:2512.00908},
  year={2025}
}
\clearpage
\appendix
\section{More about the Design of
\texorpdfstring{$\gamma_{i,t}^{+}$}{gamma (i,t) pos}}
\label{section:app:proof}

\textbf{Derivation of Equation~\ref{eq:G}.}
For a fixed state $s_{i,t}$, let $\mathbf{z}_{i,t}$ be the logit vector and
$\mathbf{e}_{y_{i,t}}$ be the one-hot target.
At each decoding step, an LLM maps logits to the next-token distribution through a softmax layer.
For any vocabulary token $v\in\mathcal{V}$, the softmax probability is

{\small
\begin{equation}
    \pi_\theta(v\mid s_{i,t})
    =
    \frac{\exp(z_v)}{\sum_{u\in\mathcal{V}}\exp(z_u)}.
\end{equation}
}
Therefore, for the sampled token $y_{i,t}$,

{\small
\begin{equation}
\begin{aligned}
    \log \pi_\theta(y_{i,t}\mid s_{i,t})
    &=
    z_{y_{i,t}}
    -
    \log\sum_{u\in\mathcal{V}}\exp(z_u).
\end{aligned}
\end{equation}
}
Taking the derivative with respect to each logit dimension $z_v$ gives

{\small
\begin{equation}
    \frac{\partial \log \pi_\theta(y_{i,t}\mid s_{i,t})}{\partial z_v}
    =
    \mathbf{1}[v=y_{i,t}]
    -
    \pi_\theta(v\mid s_{i,t}).
\end{equation}
}
Equivalently, the sampled-token dimension has gradient

{\small
\begin{equation}
    1-\pi_\theta(y_{i,t}\mid s_{i,t}),
\end{equation}
}
while every alternative token $v\neq y_{i,t}$ has gradient

{\small
\begin{equation}
    -\pi_\theta(v\mid s_{i,t}).
\end{equation}
}
Putting all dimensions together yields

{\small
\begin{equation}
    \nabla_{\mathbf{z}_{i,t}}\log \pi_\theta(y_{i,t}\mid s_{i,t})
    =
    \mathbf{e}_{y_{i,t}}-\pi_\theta(\cdot\mid s_{i,t})
    :=
    \mathbf{g}_{i,t},
\end{equation}
}
which gives Eq.~\ref{eq:G}.

This appendix provides more details about how we design Equation-(\ref{eq:gamma_pos}):

{\small
\[
    \gamma^+_{i,t} 
    = (1 - \alpha)\,\gamma_{i,t} 
    \;+\; \frac{\alpha}{N} \sum_{n=1}^N
    \mathrm{sg}\!\left(\frac{\gamma_{i,t}}{\gamma^{(n)}_{i,t}}\right)\, \gamma^{(n)}_{i,t},
\]
}
Based on Eq.~\ref{eq:G_for_positive}, the top-N smoothed target is

{\small
\[
\begin{aligned}
        \tilde{\mathbf{e}}_{y_{i,t}}
    =
    (1-\alpha)\mathbf{e}_{y_{i,t}}
    +
    \frac{\alpha}{N}\sum_{n=1}^N \mathbf{e}_{v_{i,t}^{(n)}} ,
\end{aligned}
\]
}
and its induced logit-gradient direction is

{\small
\begin{equation}
\begin{aligned}
    \tilde{\mathbf{g}}_{i,t}
    &=
    \tilde{\mathbf{e}}_{y_{i,t}}-\pi_\theta(\cdot\mid s_{i,t}) \\
    &=
    (1-\alpha)\left(\mathbf{e}_{y_{i,t}}-\pi_\theta(\cdot\mid s_{i,t})\right)
    +  \\
    &\frac{\alpha}{N}\sum_{n=1}^N
    \left(\mathbf{e}_{v_{i,t}^{(n)}}-\pi_\theta(\cdot\mid s_{i,t})\right) \\
    &=
    (1-\alpha)\mathbf{g}_{i,t}
    +
    \frac{\alpha}{N}\sum_{n=1}^N \mathbf{g}^{(n)}_{i,t},
    \label{eq:app:positive_gradient_decomp}
\end{aligned}
\end{equation}
}
where
$\mathbf{g}^{(n)}_{i,t}:=
\nabla_{\mathbf{z}_{i,t}}\log\pi_\theta(v^{(n)}_{i,t}\mid s_{i,t})$
is the logit-gradient direction for the rank-$n$ candidate.

Following the gradient-ratio decomposition used in Eq.~\ref{eq:gradient_ratio}, the unclipped GRPO
term gives

{\small
\[
\begin{aligned}
  \nabla_\theta \left(A_i\gamma_{i,t}\right)
  =&
  A_i\cdot\mathrm{sg}(\gamma_{i,t})\cdot \\
  &\underbrace{\nabla_{\mathbf{z}_{i,t}}\log\pi_\theta(y_{i,t}\mid s_{i,t})}_{\mathbf{g}_{i,t}}
  \nabla_\theta \mathbf{z}_{i,t}\,\mathbf{g}_{i,t}.
\end{aligned}
\]
}
Replacing $\mathbf{g}_{i,t}$ by the smoothed direction in
Eq.~\ref{eq:app:positive_gradient_decomp} yields

{\small
\begin{equation}
\begin{aligned}
    A_i\cdot \mathrm{sg}(\gamma_{i,t})\cdot
    \nabla_\theta\mathbf{z}_{i,t}&\tilde{\mathbf{g}}_{i,t}
    =
    (1-\alpha)
    A_i\cdot \mathrm{sg}(\gamma_{i,t})\cdot
    \nabla_\theta\mathbf{z}_{i,t}\mathbf{g}_{i,t} \\
    &\quad+
    \frac{\alpha}{N}\sum_{n=1}^N
    A_i\cdot \mathrm{sg}(\gamma_{i,t})\cdot
    \nabla_\theta\mathbf{z}_{i,t}\mathbf{g}^{(n)}_{i,t}.
    \label{eq:app:akg}
\end{aligned}
\end{equation}
}
The first term corresponds to the original sampled-token ratio $\gamma_{i,t}$.
For the rank-$n$ candidate, define

{\small
\begin{equation}
    \gamma^{(n)}_{i,t}
    =
    \frac{\pi_\theta(v^{(n)}_{i,t}\mid s_{i,t})}
    {\pi_{\theta_{\mathrm{old}}}(v^{(n)}_{i,t}\mid s_{i,t})}.
\end{equation}
}
The naive mixture

{\small
\begin{equation}
    (1-\alpha)\gamma_{i,t}
    +
    \frac{\alpha}{N}\sum_{n=1}^N\gamma^{(n)}_{i,t}
\end{equation}
}
would use different importance coefficients for different rank-$n$ candidates.
We therefore multiply each auxiliary ratio by the stop-gradient correction
$\mathrm{sg}(\gamma_{i,t}/\gamma^{(n)}_{i,t})$, so that its backward coefficient matches
$\mathrm{sg}(\gamma_{i,t})$. This design is only one line of code, using the \texttt{.detach()} in \texttt{Pytorch}.
This gives Eq.~\ref{eq:gamma_pos} and induces the smoothed gradient direction in
Eq.~\ref{eq:app:positive_gradient_decomp}.

\section{Proof for Section 3}

\subsection{Squeeze Effect of Negative Updates}
\label{app:proof_squeeze_effect}

\begin{proof}[Proof of Lemma~\ref{lem:squeeze_effect}]
Fix a state $s_{i,t}$ and write $p_v=\pi_\theta(v\mid s_{i,t})$.
Let $z_v$ be the logit of token $v$, so that
\begin{equation}
    p_v
    =
    \frac{\exp(z_v)}{\sum_{u\in\mathcal{V}}\exp(z_u)}.
\end{equation}
For the sampled token $y_{i,t}$, its probability is
\begin{equation}
    p_{y_{i,t}}
    =
    \frac{\exp(z_{y_{i,t}})}{\sum_{u\in\mathcal{V}}\exp(z_u)}.
\end{equation}
For the sampled token itself, the softmax derivative is
\begin{equation}
    \frac{\partial p_{y_{i,t}}}{\partial z_{y_{i,t}}}
    =
    p_{y_{i,t}}\left(1-p_{y_{i,t}}\right).
\end{equation}
For any alternative token $v\neq y_{i,t}$,
\begin{equation}
    \frac{\partial p_{y_{i,t}}}{\partial z_v}
    =
    -p_{y_{i,t}}p_v,
    \qquad v\neq y_{i,t}.
\end{equation}
For the unclipped GRPO term $A_i\gamma_{i,t}$, the denominator of $\gamma_{i,t}$ is fixed with
respect to the current policy, so $\gamma_{i,t}\propto p_{y_{i,t}}$.
When $A_i<0$, gradient ascent on $A_i\gamma_{i,t}$ moves in the direction of decreasing
$p_{y_{i,t}}$.
For an alternative token $v$, the corresponding first-order logit increase is therefore proportional
to $p_{y_{i,t}}p_v$:
\begin{equation}
    \Delta z_v \propto p_{y_{i,t}}p_v,
    \qquad v\neq y_{i,t}.
\end{equation}
Thus, among alternative tokens, the logit increase induced by the negative update is proportional to
the current probability $p_v$ up to the shared factor $p_{y_{i,t}}$.
If $y_{i,t}\neq v_{i,t}^{(1)}$, where $v_{i,t}^{(1)}$ denotes the rank-$1$ candidate under $\pi_\theta(\cdot\mid s_{i,t})$, then $v_{i,t}^{(1)}$ has the largest probability among alternative tokens and receives the largest logit increase.
\end{proof}

\subsection{Local Diversity under Concentration}
\label{app:proof_local_diversity}

\begin{proof}[Proof of Proposition~\ref{prop:local_diversity}]
Let $I_v$ indicate whether token $v$ appears at least once among the $K$ independent next-token
samples $\tilde{y}_{1,t},\ldots,\tilde{y}_{K,t}$:
\begin{equation}
    I_v
    =
    \mathbf{1}
    \left[
    \exists k\in\{1,\ldots,K\}: \tilde{y}_{k,t}=v
    \right].
\end{equation}
The number of distinct sampled tokens is
\begin{equation}
    \left|\{\tilde{y}_{1,t},\ldots,\tilde{y}_{K,t}\}\right|
    =
    \sum_{v\in\mathcal{V}} I_v.
\end{equation}
By linearity of expectation,
\begin{equation}
    \mathbb{E}
    \left[
    \left|\{\tilde{y}_{1,t},\ldots,\tilde{y}_{K,t}\}\right|
    \right]
    =
    \sum_{v\in\mathcal{V}}\Pr(I_v=1).
\end{equation}
Token $v$ is absent from all $K$ samples with probability $(1-p_v)^K$, so
\begin{equation}
    \Pr(I_v=1)=1-(1-p_v)^K.
\end{equation}

Let $v^{(1)}\in\arg\max_{v\in\mathcal{V}}p_v$. The contribution of $v^{(1)}$ is $1-(1-p_{v^{(1)}})^K\to1$ as $p_{v^{(1)}}\to1$.
For the remaining tokens, using $1-(1-a)^K\le Ka$ for $a\in[0,1]$,

{\small
\begin{equation}
  \sum_{v\neq v^{(1)}}
  \left(1-(1-p_v)^K\right)
  \le
  K\sum_{v\neq v^{(1)}}p_v
  =
  K(1-p_{v^{(1)}})
  \to 0.
\end{equation}
}
Therefore, the expectation in Proposition~\ref{prop:local_diversity} converges to $1$ for fixed
$K$.
\end{proof}

\section{A More Complete Related Work}
\subsection{Reinforcement Learning with Verifiable Rewards in LLMs}
Reinforcement learning with verifiable rewards (RLVR) from large language models (LLMs) has demonstrated significant potential \cite{deepseekai2025deepseekr1incentivizingreasoningcapability,zeng2025simplerlzooinvestigatingtamingzero,he2025skywork}, especially when directly applied to a base model using GRPO \cite{shao2024deepseekmath} for RL training. This approach has notably enhanced the base model's performance, particularly in improving its reasoning abilities for mathematical and coding tasks. Subsequent works have focused on improving GRPO to further enhance the algorithm’s performance.
For instance, DAPO \cite{yu2025dapo} adjusts GRPO's clipping thresholds and removes KL regularization to encourage larger updates in correct answer.  Dr.GRPO \cite{liu2025understanding} eliminates the normalization term when computing advantages to prevent length bias. GSPO \cite{zheng2025group} modifies the importance sampling from the token level to the sequence level, which proves to be more stable in training mixture-of-experts (MoE) models. Lopti \cite{yang2025not} identifies that low-probability tokens disproportionately influence GRPO training and accordingly reweights the advantage by token probability. These modifications have contributed to improvements in the model's pass@1 performance, but they have not specifically addressed pass@K performance, which relates to the model’s exploration ability.

\subsection{Effective Exploration for RLVR in LLMs }

A central challenge in RLVR tasks lies in moving beyond the exploitation of a pretrained model's implicit knowledge to actively exploring diverse reasoning paths. Current methods tend to converge on a limited set of solutions, as evidenced by poor performance on the pass@K metric, which evaluates the coverage of multiple reasoning paths and thus reflects exploration effectiveness \cite{yue2025does,wu2025invisible}. To address this exploration deficit, existing methods can be roughly grouped by the level at which they intervene.

\emph{Response-level} methods intervene at the level of whole trajectories. A representative example is pass@K training and related reward-shaping methods, which optimize the response distribution by directly reweighting sampled solutions \cite{walder2025pass, chen2025pass, li2026setpo, tajwar2026maximum, zhang2025rspo, li2025jointly, liu2026past, gai2025differential, dai2025cde}. These methods improve exploration in a trajectory-level sense, but they treat the model as a black box and do not explain which local candidates should receive probability mass.

\emph{Token-level} methods further modify the local learning signal. Some methods use entropy as a proxy for exploration \cite{cui2025entropy,cheng2025reasoning, wang2025beyond, hou2025advancing, hu2025openreasonerzeroopensourceapproach, jiang2025rethinking}. Others use proxy signals such as confidence or hidden-state similarity to reweight token-level advantages \cite{deng2025token, lin2026resrl, yangdcpo}. However, these proxies remain coarse summaries of exploration behavior and do not provide fine-grained control over which candidates should retain probability mass.
A few recent methods combine token-level and response-level objectives \cite{yangdcpo, wandsdr, dai2025cde, song2025outcome}.

Finally, \emph{data-centric} methods expand the training distribution to expose the model to more diverse reasoning environments. One line of work
uses off-policy data from more capable models to broaden the model's knowledge and promote solution diversity \cite{dong2025rl, li2025questa}.
Additional strategies include generating varied responses for challenging samples \cite{yang2025depth} or paraphrasing questions to stimulate
different reasoning trajectories for the same problem \cite{liang2025beyond}. These methods are effective to some extent, but they require either additional expert data or extra computation to generate diverse training data, rather than directly controlling how probability mass is allocated during training.

Another relevant line of work decomposes RLVR updates into positive and negative components, arguing that positive reinforcement sharpens the distribution while negative reinforcement enhances exploration \citep{zhu2025surprising}. However, as we analyze in Section~\ref{sec:negative_updates}, negative gradients do not uniformly promote exploration: when applied to low-probability tokens, they can in fact sharpen the distribution and induce a squeezing effect, which was not captured  in prior analyses. Moreover, as shown in Sections~\ref{sec:math} and \ref{sec:logical_code}, negative-only reinforcement improves exploration but at the cost of exploitation. It increases pass@K but consistently decreases pass@1. In particular, on tasks where the model must first consolidate exploitation before exploration becomes beneficial, strong negative reinforcement disrupts early pattern learning, ultimately reducing both pass@1 and pass@K. Our work addresses these limitations by providing a token-level, learning-dynamics–based perspective that explains these behaviors and motivates our proposed approach.
\subsection{Analysis of Learning Dynamics in LLMs}
Analyzing the learning dynamics of deep neural networks provides valuable insight into how training shapes model behavior \cite{saxe2013exact, tachet2018learning,advani2020high}. This analytical perspective has recently been extended to Large Language Models (LLMs), where prior work has widely examined the dynamics of supervised fine-tuning (SFT) \cite{kang2024learning,chu2025sft}, off-policy preference optimization methods such as DPO \cite{razin2025unintentional,pal2024smaug}, or both \cite{ren2024learning,  ren2025thesis}.

Several recent studies have begun exploring the learning dynamics of on-policy RL.
\cite{cui2025entropy} adopt entropy-based metrics to track model changes during training. However, such metrics provide only an indirect signal by averaging over the entire vocabulary, thereby failing to capture meaningful shifts among high-probability candidates.
In contrast, \cite{deng2025effect} examine probability shifts induced by individual gradient updates to analyze inter-sample effects. While these analyses offer valuable fine-grained insights into probability changes, they fail to capture the cumulative evolution of the model’s policy.
To overcome these limitations, we propose a top-N probability dynamics framework that directly tracks how probability mass redistributes among the most likely candidates throughout training. This approach provides a scalable and interpretable lens for understanding how on-policy RL shapes model behavior.

\section{Additional Analysis of Distribution Changes During Training}


\begin{figure*}[t!]
  \centering
  \setlength{\abovecaptionskip}{4pt}
  \setlength{\belowcaptionskip}{-6pt}
  \includegraphics[width=0.99\textwidth]{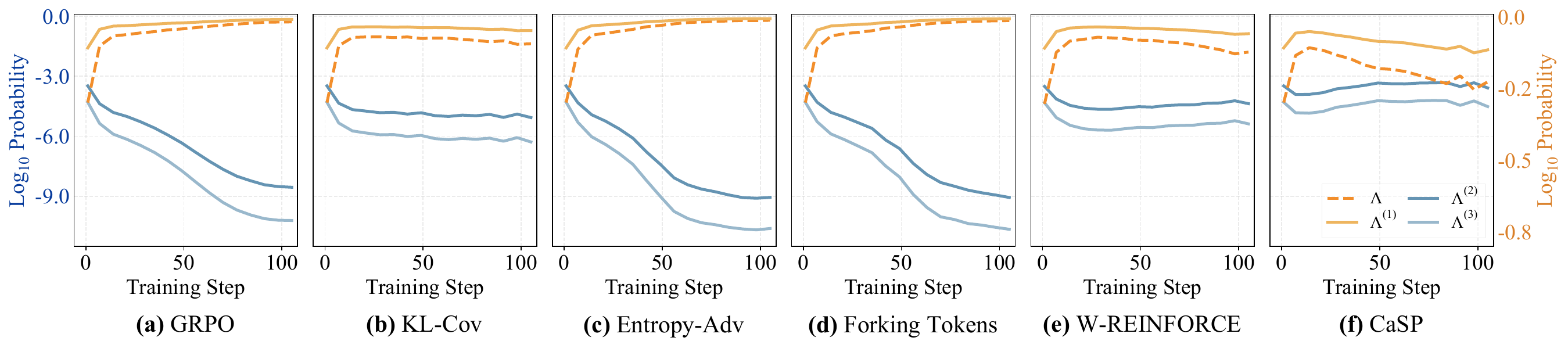}
\caption{ Comparison of CaSP with GRPO, KL-Cov, ``forking" tokens, W-REINFORCE and Entropy-Adv on Qwen2.5-Math-7B. CaSP effectively controls probability concentration on the $\Lambda^{(1)}$ while preserving diversity among $\Lambda^{(2)}$
  and $\Lambda^{(3)}$.}
  \label{fig:analysis_simko_appendix}
\end{figure*}
\subsection{Effects of CaSP on Training Dynamics}
\label{appendix:train_dynamic}

We analyze the training dynamics of CaSP in comparison with GRPO, KL-Cov, ``forking'' tokens, W-REINFORCE and Entropy-Adv.  Figure~\ref{fig:analysis_simko_appendix} present the changes of top-N log-probabilities ($\Lambda^{(1)}$, $\Lambda^{(2)}$, and $\Lambda^{(3)}$) across training steps.

As can be seen, GRPO leads to severe over-concentration: $\Lambda^{(1)}_{\text{GRPO}}$ increases to nearly 0, while $\Lambda^{(2)}_{\text{GRPO}}$ and $\Lambda^{(3)}_{\text{GRPO}}$ sharply drop below -8 and -10, respectively. This indicates that nearly all probability mass collapses onto the top-1 token. KL-Cov exhibits a moderate concentration effect due to the KL penalty, while Entropy-Adv collapses even more rapidly, likely because of its stronger emphasis on high-entropy tokens. Training only on forking tokens further exacerbates probability over-concentration compared to GRPO, as it selectively increases the probability of high-entropy tokens. W-REINFORCE alleviates part of the over-concentration, but its sequence-level penalization of negative samples induces the squeezing effect in Section~\ref{sec:negative_updates}, leaving it still prone to probability concentration. In contrast, CaSP achieves the most effective deconcentration among all methods. This is evidenced by a lower $\Lambda^{(1)}_{\text{CaSP}}$ and higher $\Lambda^{(2)}_{\text{CaSP}}$ and $\Lambda^{(3)}_{\text{CaSP}}$. These results suggest that CaSP effectively mitigates probability mass collapse and can potentially encourages exploration during training.

\begin{figure*}[t]
    \centering
      \setlength{\abovecaptionskip}{3pt}
  \setlength{\belowcaptionskip}{-6pt}
    \includegraphics[width=0.9\linewidth]{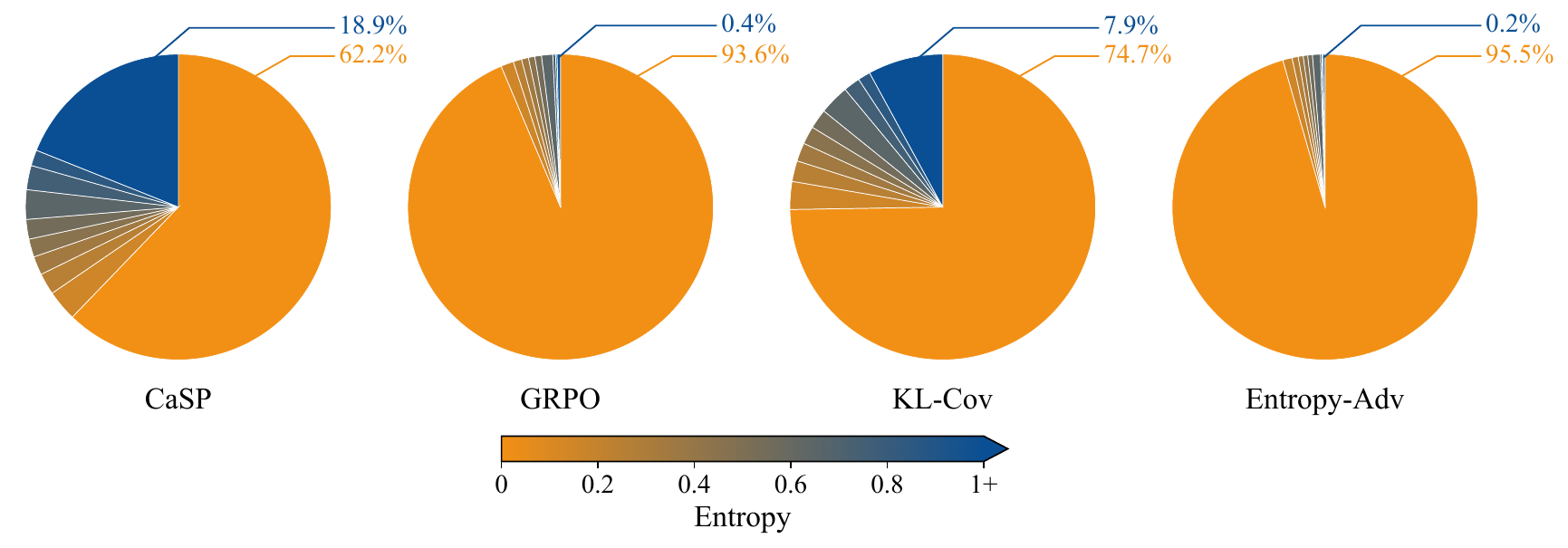}
    \caption{ Token-level entropy distributions from the Qwen2.5-Math-7B backbone trained with CaSP, GRPO, KL-Cov, and Entropy-Adv, demonstrating CaSP's ability to maintain the entropy of the ``forking" tokens.}
    \label{fig:entropy_distribution}
\end{figure*}

To further validate this, we visualize the histogram of token-level entropy in Figure \ref{fig:entropy_distribution}. GRPO drives most tokens toward near-zero entropy. CaSP, however, can preserve token entropy, particularly at ``forking'' points, where high entropy is desirable for exploration. This preservation of entropy further  confirms CaSP's role in promoting exploration.

\begin{figure*}[!ht]
    \centering
    \setlength{\abovecaptionskip}{2pt}
\setlength{\belowcaptionskip}{-8pt}    \includegraphics[width=0.8\linewidth]{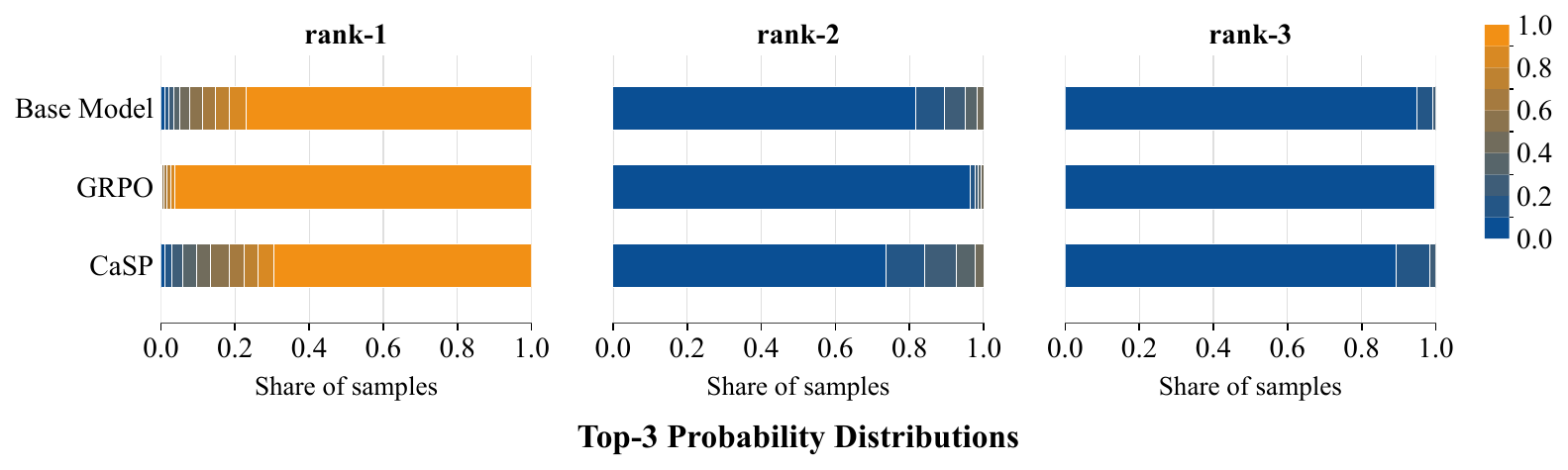}
\caption{Probability-bin distributions for the top-3 next-token candidates. Across methods, the rank-1 candidate occupies substantially higher probability bins, whereas the rank-2 and rank-3 candidates are concentrated in the lowest bin. This pattern suggests that tracking the top-3 candidates captures the dominant changes in the model's next-token distribution.}
    \label{fig:top3_probability}
\end{figure*}

\vspace{-1mm}
\subsection{Top-3 Candidates Probabilities Distribution}

Figure~\ref{fig:top3_probability} illustrates the probability-bin distributions of the top-3 candidates. The rank-$1$ candidate is concentrated in high-probability bins for all models, with the GRPO-trained model showing the strongest concentration. In contrast, the rank-$2$ and rank-$3$ candidates are mostly assigned to the lowest probability bin, and the rank-$3$ candidate is below 0.05 in more than 95\% of cases across methods. This sharp decay from rank-$1$ to rank-$3$ indicates that most of the relevant probability mass is captured by the leading candidates, supporting our use of top-N candidate statistics as a tractable proxy for monitoring changes in the full next-token distribution.

\vspace{-1mm}
\section{Implementation Details}

\vspace{-1mm}
\subsection{Training Details}
\label{appendix:training_detail}

All experiments are conducted on 8 H100 GPUs. For the math tasks, training takes approximately 120 H100 GPU hours for the 7B models, while Qwen2.5-32B requires nearly 192 H200 GPU hours. For the logic tasks, training takes about 120 H100 GPU hours. For the coding tasks, training takes about 160 H100 GPU hours. We use verl v0.2.0 for training.
\paragraph{Math Tasks.}
For math tasks, all models are trained with a learning rate of $10^{-6}$, a global batch size of 1024, and a PPO mini-batch size of 256. For each input problem, we sample 8 responses with temperature 1.0. The maximum prompt length is 1024. The maximum response length is 3072 for Qwen2.5-Math-7B and Llama3.2-3B-Instruct, and 8192 for Qwen2.5-32B. For the entropy-loss baseline, we set the coefficient to 0.01. For CaSP, we set $\alpha=0.01$ and define the entropy threshold $\tau_q$ as the $q$-quantile of the token-level entropy distribution, such that a fraction $q$ of tokens in each response have entropy lower than $\tau_q$. Unless otherwise specified, we use $\tau_{0.8}$. We set $\lambda_{\mathrm{top1}}=1.1$ and $N=3$ for all math models.

\paragraph{Logic Tasks.}
For logic tasks, all models are trained with a learning rate of $10^{-6}$, a global batch size of 128, and a PPO mini-batch size of 32. For each input problem, we sample 16 responses with temperature 1.0. The maximum prompt length is 2048, and the maximum response length is 8192. We use dynamic sampling and set the upper clipping threshold to 0.28 for all methods. For CaSP, we apply a 50-step GRPO warm-up and set $\alpha=0.005$, $ \lambda_{\mathrm{top1}}=1.05$, and $N=3$.

\paragraph{Coding Tasks.}
For coding tasks, all models are trained with a learning rate of $10^{-6}$, a global batch size of 128, and a PPO mini-batch size of 32. For each input problem, we sample 8 responses with temperature 0.6. The maximum prompt length is 2048, and the maximum response length is 8192. For CaSP, we set $\alpha=0.05$, $\lambda_{\mathrm{top1}} =1.1$, and $N=3$.

\vspace{-1mm}
\subsection{Evaluation Details}
\label{appendix:evaluation_detail}

\paragraph{Math Tasks.}
For math evaluation, we report pass@1 and pass@K. We use temperature 0.6 and top-$p=0.95$ for sampling. The maximum generation length is 4096 for Qwen2.5-Math-7B and Llama3.2-3B-Instruct, and 8192 for Qwen2.5-32B. To reduce variance on small datasets such as AIME and AMC, we sample $n=300$ responses per problem. For other math benchmarks, we use $n=256$, except for Qwen2.5-32B where we use $n=128$.

\paragraph{Logic Tasks.}
For logic evaluation, we sample $n=128$ responses per problem with temperature 0.7, top-$p=0.95$, and a maximum generation length of 8192. We report pass@1 and pass@K using the same evaluation protocol across all compared methods.

\paragraph{Coding Tasks.}
For coding evaluation, we sample $n=16$ responses per problem with temperature 0.7, top-$p=0.95$, and a maximum generation length of 8192. We report pass@1 and pass@K on the coding benchmarks.

\vspace{-1mm}
\subsection{PseudoCode}
\label{appendix:pseudocode}
We present the pseudocode in Figure~\ref{code}.

\vspace{-1mm}
\subsection{Artifacts and Licenses}
\label{appendix:artifacts_licenses}

We use publicly available models, datasets, and software artifacts in this work. The pretrained models include
Qwen2.5-Math-7B, Qwen2.5-32B, Llama3.2-3B-Instruct, and DeepSeek-R1-Distill-Qwen-7B, each subject to the license
and terms released by its original provider. The training and evaluation datasets, including MATH, GSM8K,
SynLogic, BBH, LiveCodeBench, HumanEval+, AIME, AMC, Minerva Math, and OlympiadBench, are used according to their
respective public licenses or benchmark usage terms.

\subsection{Artifact Use and Data Considerations}
  \vspace{-1mm}
  \label{appendix:artifact_data}
 \setlength{\abovecaptionskip}{2pt}
    \setlength{\belowcaptionskip}{-8pt}
We use publicly available models and benchmarks only for their intended research and evaluation purposes. The datasets are public research benchmarks, and we do not collect new user data or include personally identifying information.

\begin{figure}[!t]
\centering
\setlength{\abovecaptionskip}{1pt}
  \setlength{\belowcaptionskip}{-3pt}
\begin{tcolorbox}[
    colframe=gray,       %
    colback=white,       %
    boxrule=0.5pt,       %
    arc=2pt,             %
    left=0pt, right=4pt, top=0pt, bottom=4pt, %
    width=\linewidth,    %
    enhanced,
]

\begin{normalcode}
def compute_policy_loss():
    ratio = torch.exp( log_prob - old_log_prob )
\end{normalcode}

\begin{diffaddcode}
+    # 1. Identify forking tokens
+    w = (entropy > percentile(entropy, $\tau$))
+    # 2. Using top-N ratio
+    topn_ratio = torch.exp(topn_log_probs - old_topn_log_probs)
+    topn_ratio = ((ratio.detach() / topn_ratio.detach())*topn_ratio).sum(dim=-1)
+    ratio = torch.where(advantage > 0, (1-$\alpha$*w)*ratio + ($\alpha$*w/K)*topn_ratio, ratio)
+    # 3. Apply a strong penalty to top1 negative tokens
+    mask = (advantage < 0) & is_top1 & w
+    ratio[mask] *= $\lambda$
\end{diffaddcode}
\begin{normalcode}
    pg_losses = -advantage * ratio
    # ...clip and compute loss
\end{normalcode}
\end{tcolorbox}
\caption{ The pseudo-code of the policy loss computation with CaSP.  CaSP only requires modifying a few lines from a standard policy gradient implementation.}
\label{code}
\end{figure}

\section{More Experiment Result}

\subsection{Detailed Ablation Results}
\label{appendix:ablation result}
In this section, we present more detailed ablation results. As shown in Table~\ref{ablation-result}, applying CaSP exclusively to either correct or incorrect responses leads to a drop in pass@$K$ performance. This highlights the importance of both CaSP designs, as applying both yields the best results.

\begin{table*}[!ht]

\centering
\caption{Ablations on $\alpha$, $\tau$, $k$, $\gamma^+$, and  $\gamma^-$. Pass@1/Pass@256 scores are evaluated using Qwen2.5-Math-7B.}
\label{ablation-result}
\renewcommand{\arraystretch}{1.25}
\setlength{\tabcolsep}{8.5pt}      
\footnotesize
\begin{tabular}{lccccccc}
\textbf{Method} & \textbf{AIME24} & \textbf{AIME25} & \textbf{AMC} & \textbf{MATH500}& \textbf{Minerva} & \textbf{Olympiad} & \textbf{Avg.} \\
\hline
Base Model & 13.2 / 66.0 & 5.4 / 51.8 & 38.2 / 98.5 & 55.8 / 96.0 & 16.5 / 68.8 & 25.6 / 77.0 & 25.8 / 76.4 \\
GRPO & 28.1 / 72.3 & 11.5 / 52.1 & 61.2 / 97.1 & 76.6 / 96.2 & 33.4 / 64.0 & 39.1 / 74.7 & 41.7 / 76.1 \\ \hline
\(\alpha\)=0.02 & 31.1 / 79.9 & \textbf{13.3} / 58.9 & 62.6 / 99.3 & \textbf{77.6} / \textbf{97.0} & 35.2 / 66.2 & 39.4 / 76.9 & 43.2 / 79.7 \\
\(\alpha\)=0.03 & 30.9 / 72.7 & 12.4 / 64.6 & 62.4 / 97.5 & 77.2 / 97.6 & 34.9 / 66.9 & 38.9 / 76.9 & 42.8 / 79.4 \\
\(\alpha\)=0.05 & 30.8 / 78.7 & 12.2 / \textbf{67.8} & 63.0 / 97.5 & 77.2 / 96.8 & 34.8 / 65.1 & 39.3 / 76.3 & 42.9 / 80.4 \\
\(\alpha\)=0.1  & 29.1 / 75.5 & 12.4 / 65.0 & 61.8 / \textbf{99.9} & 77.2 / 96.8 & 35.2 / 67.6 & 38.9 / \textbf{77.8} & 42.4 / 80.4 \\ \hline
\(\tau_0\)   & 10.7 / 74.1 & 6.2 / 54.0 & 51.7 / 96.7 & 70.5 / 95.8 & 32.2 / 67.3 & 33.5 / 74.5 & 34.1 / 77.1 \\
\(\tau_{0.4}\)   & 20.8 / 70.9 & 7.2 / 57.5 & 58.3 / 94.6 & 74.9 / 95.2 & 33.8 / 68.0 & 36.4 /  75.0 & 38.6 / 76.9 \\
\(\tau_{0.6}\)   & 27.5 / 77.6 & 10.0 / 56.0 & 62.2 / \textbf{99.9} & 76.5 / 96.8 & \textbf{35.3} / \textbf{68.8} & 38.2 / 76.4 & 41.6 / 79.3 \\
\hline
$K=1$ & 29.5 / 83.7 & 12.2 / 55.1 & 62.1 / 97.1 & 77.1 / 96.8 & 35.1 / 65.8 & 39.5 / 75.9 & 42.6 / 79.1 \\
$K=2$ & 30.3 / 78.4 & 12.2 / 57.9 & 62.5 / 99.6& 77.2 / 96.4 & 34.9 / 65.8 & 39.5 / 76.9 & 42.8 / 79.2 \\
$K=4$ & 32.8 / \textbf{84.6} & 12.5 / 54.6 & 62.6 / 97.5 & 77.7 / 97.2 & 35.3 / 68.4 & 39.2 / 78.4 & 43.4 / 80.1 \\
$K=5$ & 31.3 / 78.5 & 11.7 / 58.0 & 62.2 / 99.6 & 77.5 / 96.2 & 35.7 / 68.0 & 39.0 / 76.3 & 42.9 / 79.4 \\
\hline
w/o \(\gamma^-\)  & 31.5 / 80.7 & 11.5 / 57.9 & \textbf{62.7} / 97.4 & 77.1 / 96.2 & 34.1 / 65.8 & 39.0 / 75.6 & 42.8 / 78.9 \\
w/o \(\gamma^+\) & 30.4 / 75.2 & 12.7 / 64.9 & 62.3 / 99.6 & 77.4 / 96.4 & 34.7 / 65.8 & 39.5 / 77.3 & 42.8 / 79.9 \\ \hline
\rowcolor{Gray}
CaSP & \textbf{32.8} / 78.0 & 12.9 / 64.6 & 62.4 / 97.5 & \textbf{77.6} / 96.8 & 35.0 / 68.4 & \textbf{39.8} / \textbf{77.8} & \textbf{43.4} / \textbf{80.5} \\
\end{tabular}
\end{table*}

\begin{table*}[!ht]
\centering
\renewcommand{\arraystretch}{1.25}
\caption{Pass@1 / Pass@256 results for  Qwen2.5-Math-7B across different temperatures on MATH500, AIME 2024/25, Minerva\_math, Olympiadbench, and AMC23 Datasets.}
\label{math-temperature-appendix}
\footnotesize
\begin{tabular}{lcccccccc}
\textbf{Method} & \textbf{Temp} & \textbf{AIME24} & \textbf{AIME25} & \textbf{AMC23} & \textbf{MATH500}& \textbf{Minerva} & \textbf{Olympiad} & \textbf{Avg.} \\
\hline
Base Model & 0.2& 13.3 / 55.0&5.3 / 41.9&40.7 / 82.5&62.8 / 93.2&14.8 / 44.1&28.2 / 66.2&27.5 / 63.8 \\
GRPO & 0.2& 27.0 / 62.7&11.8 / 53.7&64.0 / 89.6&76.3 / 90.0&34.1 / 55.9&38.6 / 66.5&42.0 / 69.7 \\
\rowcolor{Gray} CaSP & 0.2& \textbf{31.9} / \textbf{72.43}&\textbf{13.2} / \textbf{57.5}&\textbf{63.9} / \textbf{99.8}&\textbf{78.5} / \textbf{96.8}&\textbf{36.2} / \textbf{62.1}&\textbf{40.7} / \textbf{74.7}&\textbf{44.1} / \textbf{77.2} \\

\hline

Base Model& 0.4& 13.8 / 58.0& 5.5 / 50.9& 42.4 / 91.7& 61.1 / 95.4& 15.6 / 61.8& 27.6 / 73.5& 27.7 / 71.9 \\
GRPO& 0.4& 26.7 / \textbf{75.6}& 12.2 / 54.6& \textbf{63.4} / 94.5& 76.4 / 93.2& 33.5 / 61.0& 38.6 / 71.6& 41.8 / 75.1 \\
\rowcolor{Gray} CaSP& 0.4& \textbf{31.1} / 72.3& \textbf{13.0} / \textbf{64.1}& 62.9 / \textbf{97.5}& \textbf{78.1} / \textbf{97.4}& \textbf{36.1} / \textbf{66.2}& \textbf{40.6} / \textbf{76.1}& \textbf{43.6} / \textbf{78.9} \\
\hline

Base Model& 0.6& 13.2 / 66.0& 5.4 / 51.8& 38.2 / 98.5& 55.8 / 96.0& 16.5 / \textbf{68.8}& 25.6 / \textbf{77.0}& 25.8 / 76.4 \\
GRPO& 0.6& 26.5 / 75.1& 12.3 / 52.3& 62.9 / 98.9& 76.5 / 93.8& 33.9 / 63.6& 38.5 / 73.8& 41.8 / 76.3 \\
CaSP& 0.6& \textbf{30.5} / \textbf{84.0}& \textbf{12.5} / \textbf{58.5}& \textbf{63.4} / \textbf{99.9}& \textbf{77.6} / \textbf{96.8}& \textbf{36.3} / 66.9& \textbf{40.0} / 76.6& \textbf{43.4} / \textbf{80.5} \\
\hline

Base Model& 0.8& 26.2 / 75.5& 4.2 / 48.5& 34.8 / \textbf{99.9}& 50.8 / 96.8& 17.0 / \textbf{69.9}& 23.1 / \textbf{79.6}& 23.5 / 76.7 \\
GRPO& 0.8& 26.6 / \textbf{81.7}& 12.0 / \textbf{52.3}& 61.9 / 97.4& 76.5 / 95.2& 33.5 / 64.0& 38.5 / 74.5& 41.5 / 77.5 \\
\rowcolor{Gray} CaSP& 0.8& \textbf{30.1} / 78.9& \textbf{12.5} / \textbf{52.3}& \textbf{63.2} / 99.5& \textbf{77.0} / \textbf{97.2}& \textbf{35.5} / 67.6& 3\textbf{9.2} / 77.5& \textbf{42.9} / \textbf{78.8} \\
\hline

Base Model& 1.0& 9.0 / 66.0& 3.3 / 52.6& 28.8 / \textbf{99.6}& 42.8 / \textbf{97.2}& 15.3 / \textbf{71.7}& 19.0 / \textbf{80.4}& 19.7 / 77.9 \\
GRPO& 1.0& 26.2 /75.5& \textbf{11.7} / 51.8& 61.5 / 97.1& \textbf{76.4} / 95.6& 33.5 / 63.2& \textbf{38.3} / 75.7& 41.3 / 76.6 \\
\rowcolor{Gray} CaSP& 1.0& \textbf{28.9} / \textbf{75.6}& 11.0 / \textbf{64.5}& \textbf{61.9} / 97.4& 76.0 / 96.6& \textbf{34.8} / 69.5& 37.9 / 77.2& \textbf{41.8} / \textbf{80.1} \\

\end{tabular}
\end{table*}

\subsection{The Influence of Sampling Temperature}

We evaluate Qwen2.5-Math-7B and its variants trained with GRPO and CaSP under different decoding temperatures ranging from 0.2 to 1.0. The results are shown in Table~\ref{math-temperature-appendix}. Across all temperatures, CaSP consistently outperforms GRPO on both pass@1 and pass@K. 

Benefiting from the probability smoothing introduced on ``forking" tokens, CaSP maintains strong pass@256 performance even at very low temperatures. For example, at temperature 0.2, CaSP achieves 77.2\% in pass@256, compared to only 63.8\% and 69.7\% for Qwen2.5-Math-7B and GRPO, respectively. This indicates that CaSP enables the trained model to retain a broader solution paths.
At higher temperatures, all methods become more exploratory, yet CaSP consistently preserves a clear advantage. These results demonstrate that our conclusions are not sensitive to a narrow temperature choice and that CaSP is robust to this decoding hyperparameter.

\subsection{Effect of Non-Uniform Top-N Weight Allocation}
To examine whether non-uniform weight allocation within the top-N set helps, we implemented a probability-proportional softmax variant, referred to as CaSP (softmax), in place of the uniform smoothing used in the main CaSP method. The results in Table~\ref{simko-softmax}  show that this probability-aware smoothing yields a slight improvement in pass@K,indicating that allocating mass among plausible alternatives is indeed beneficial; however, it also introduces a decrease in pass@1 compare to CaSP.
We regard this as a promising avenue for further refinement. At the same time, the observation reinforces our core motivation: maintaining diversity at  “forking’’ tokens improves pass@K.

\vspace{-1mm}
\subsection{Scaling Sample K to 1024}
 We evaluate up to pass@1024 on AIME24/25 following \cite{yue2025does}. The results in Table~\ref{aime1024} show that even in cases where the base model surpasses GRPO (K = 1024), CaSP still delivers large improvements over the base model. These findings further indicate that our method enhances exploration and has strong potential for handling more challenging tasks.

\section{Qualitative Analysis of Local Candidate Support}

To qualitatively examine next-token exploration, we visualize grouped candidate matrices for two response segments in Figures~\ref{Qualitative_base1}--\ref{Qualitative_casp2}. Columns denote generated token positions. For each position, candidates are added in descending probability order until their cumulative mass reaches 0.95. Thus, fewer visible candidates indicate stronger local probability concentration. Darker cells indicate higher probability, and the highlighted path marks the sampled sequence.

The visualizations show how many local candidates are needed to cover 95\% of the next-token probability mass at each sampled position. The base model retains a small but visible set of alternatives around the sampled trajectory. After GRPO training, many positions require only one  candidate to reach the 0.95
threshold, indicating stronger probability concentration and fewer effective branches for sampling.

In contrast, CaSP keeps broader candidate support around the sampled path: several positions require additional plausible candidates to reach the 0.95 cumulative mass threshold. This suggests that CaSP preserves alternative reasoning branches while maintaining the main trajectory, supporting our claim that top-$N$ candidate support improves exploration without disrupting exploitation.
\begin{table*}[!t]
\centering
\caption{ Pass@1 / Pass@256 results for  Qwen2.5-Math-7B with the probability-proportional softmax variant of CaSP across MATH500, AIME 2024/25, Minerva\_math, Olympiadbench, and AMC23 Datasets.}
\label{simko-softmax}
\renewcommand{\arraystretch}{1.25}
\setlength{\tabcolsep}{8pt}      
\footnotesize
\begin{tabular}{lccccccc}
\textbf{Method} & \textbf{AIME24} & \textbf{AIME25} & \textbf{AMC} & \textbf{MATH500}& \textbf{Minerva} & \textbf{Olympiad} & \textbf{Avg.} \\
\hline
Base Model & 13.2 / 66.0 & 5.4 / 51.8 & 38.2 / 98.5 & 55.8 / 96.0 & 16.5 / 68.8 & 25.6 / 77.0 & 25.8 / 76.4 \\
GRPO & 28.1 / 72.3 & 11.5 / 52.1 & 61.2 / 97.1 & 76.6 / 96.2 & 33.4 / 64.0 & 39.1 / 74.7 & 41.7 / 76.1 \\ \hline
\rowcolor{Gray}
CaSP (softmax) & 29.8 / \textbf{81.6} & 12.4 / 63.2 & 61.5 / \textbf{99.6} & 77.3 / \textbf{97.2} & 34.0 / 66.9 & 39.4 / \textbf{77.8} & 42.4 / \textbf{81.1} \\
\rowcolor{Gray}
CaSP & \textbf{32.8} / 78.0 & \textbf{12.9} / \textbf{64.6} & \textbf{62.4} / 97.5 & \textbf{77.6} / 96.8 & \textbf{35.0} / 68.4 & \textbf{39.8} / \textbf{77.8} & \textbf{43.4} / 80.5 \\
\end{tabular}

\end{table*}

\begin{table*}[!ht]
\centering
\caption{ Pass@1024 results for Qwen2.5-Math-7B on AIME 2024/25 Datasets.}
\label{aime1024}
\renewcommand{\arraystretch}{1.25}
\setlength{\tabcolsep}{8pt}  
\setlength{\abovecaptionskip}{2pt}
\setlength{\belowcaptionskip}{4pt}

\footnotesize
\begin{tabular}{clccccccccccc}

\textbf{Dataset} & \textbf{Method} & 1 & 2 & 4& 8& 16 & 32 & 64 & 128 & 256 &512 &1024 \\
\hline
\multirow{3}{*}{AIME24}
 &Base Model & 13.4 &	21.3	&29.6	&36.7	&42.9	&49.1	&55.2	&60.8	&66.1	&72.0	&80.8 \\ 
  &GRPO & 27.0	&33.3	&39.2	&45.0	&51.4	&58.1	&64.6	&70.4	&74.4	&76.2	&76.7 \\ 
  \rowcolor{Gray}
 &CaSP  & \textbf{30.3} & \textbf{39.4} & \textbf{46.3} & \textbf{51.9} & \textbf{57.1} & \textbf{61.6} & \textbf{66.0} & \textbf{71.1} & \textbf{76.5} & \textbf{81.2} & \textbf{85.6}\\  
\hline
\multirow{3}{*}{AIME25}
 &Base Model & 5.6	&9.5	&14.7&	20.8&	27.0	&32.9	&38.6&	44.3&	50.5&	57.9&	65.5 \\ 
  &GRPO & 12.0	&17.3	&23.4	&29.5	&34.7&	39.2&	43.5	&47.9	&53.1&	58.3	&62.3 \\ 
  \rowcolor{Gray}
 &CaSP  & \textbf{12.9}	&\textbf{19.2}	&\textbf{25.4}	&\textbf{30.9}	&\textbf{35.8}	&\textbf{40.4}&	\textbf{44.7}&	\textbf{49.5}&	\textbf{55.6}	&\textbf{62.5}&	\textbf{68.9}\\  
\end{tabular}

\end{table*}

\begin{figure*}[t!]
  \centering
  \setlength{\abovecaptionskip}{4pt}
  \setlength{\belowcaptionskip}{0pt}
  \includegraphics[width=0.99\textwidth]{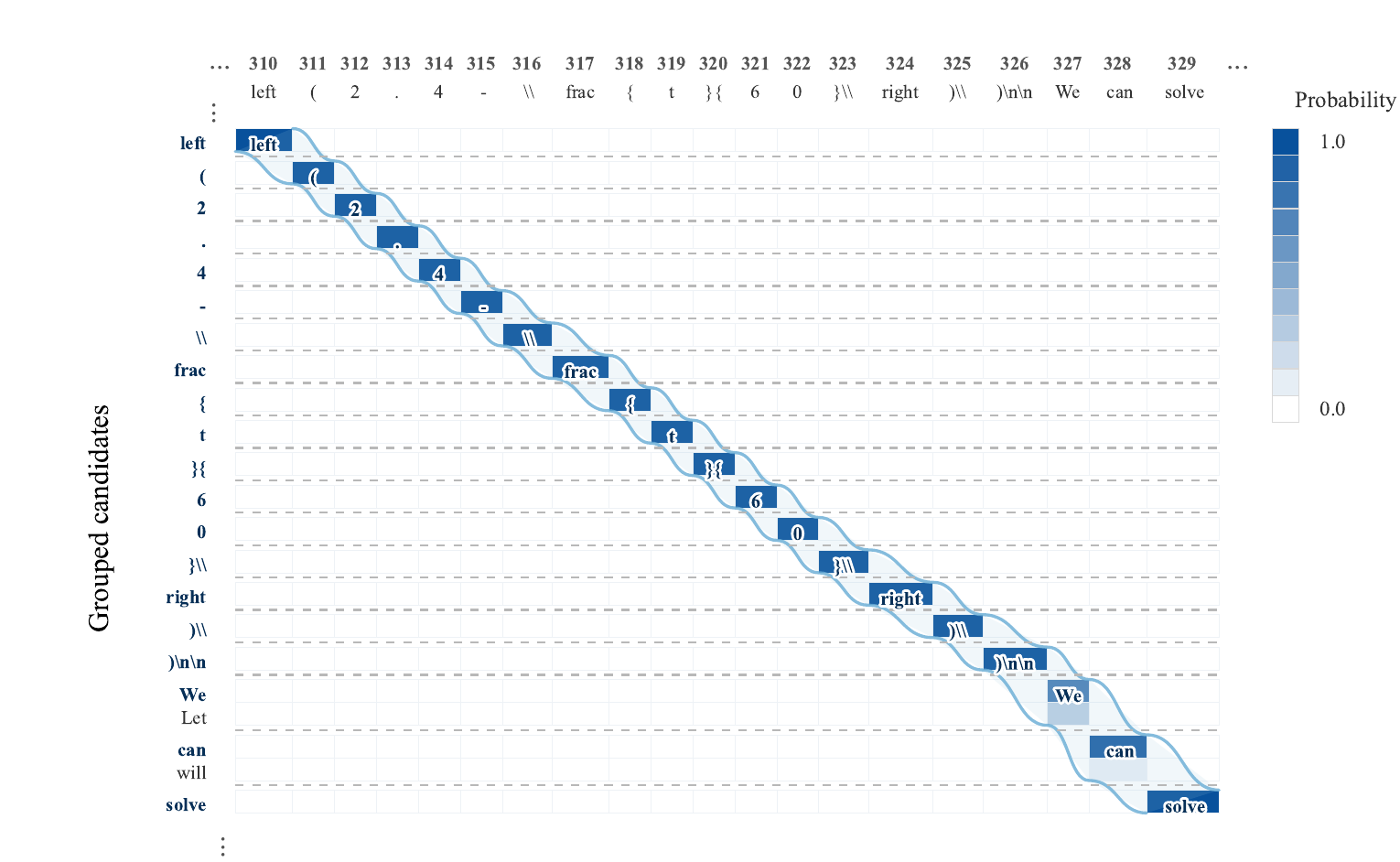}
\caption{Grouped candidate matrix for the base model on response tokens 310--329. For each position, candidate tokens are shown in descending probability order until their cumulative mass reaches 0.95, with the sampled token always included. Fewer visible candidates therefore indicate stronger local probability concentration. Darker cells indicate higher next-token probability, and the highlighted path marks the
  sampled token sequence.}
  \label{Qualitative_base1}
\end{figure*}
\begin{figure*}[t!]
  \centering
  \setlength{\abovecaptionskip}{4pt}
  \setlength{\belowcaptionskip}{0pt}
  \includegraphics[width=0.99\textwidth]{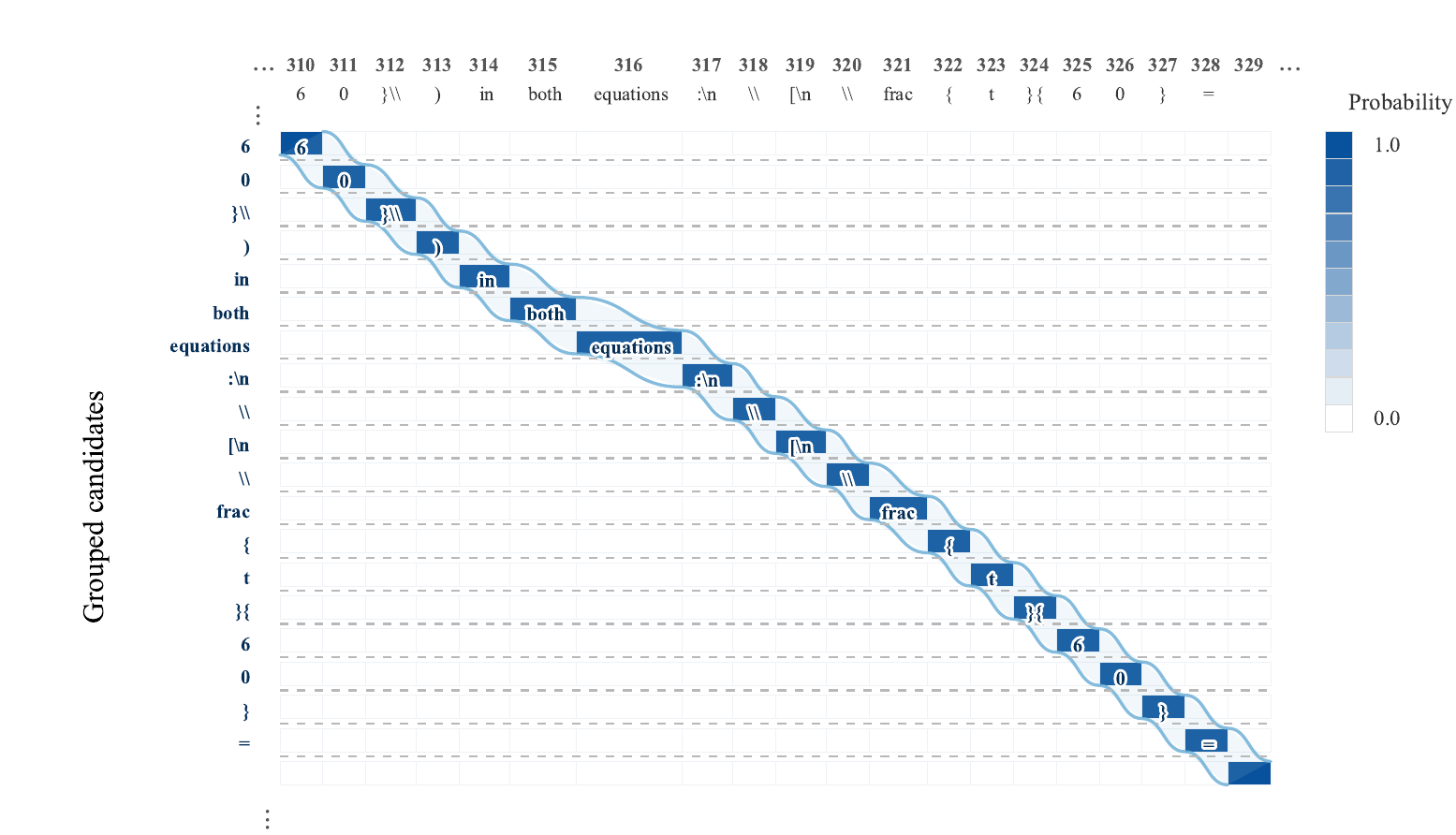}
\caption{Grouped candidate matrix for the GRPO-trained model on response tokens 310--329. Compared with the base model, GRPO concentrates probability more strongly on the sampled/rank-1 path, leaving fewer visible high-probability alternatives.}
\label{Qualitative_grpo1}
\end{figure*}
\begin{figure*}[t!]
  \centering
  \setlength{\abovecaptionskip}{4pt}
  \setlength{\belowcaptionskip}{0pt}
  \includegraphics[width=0.99\textwidth]{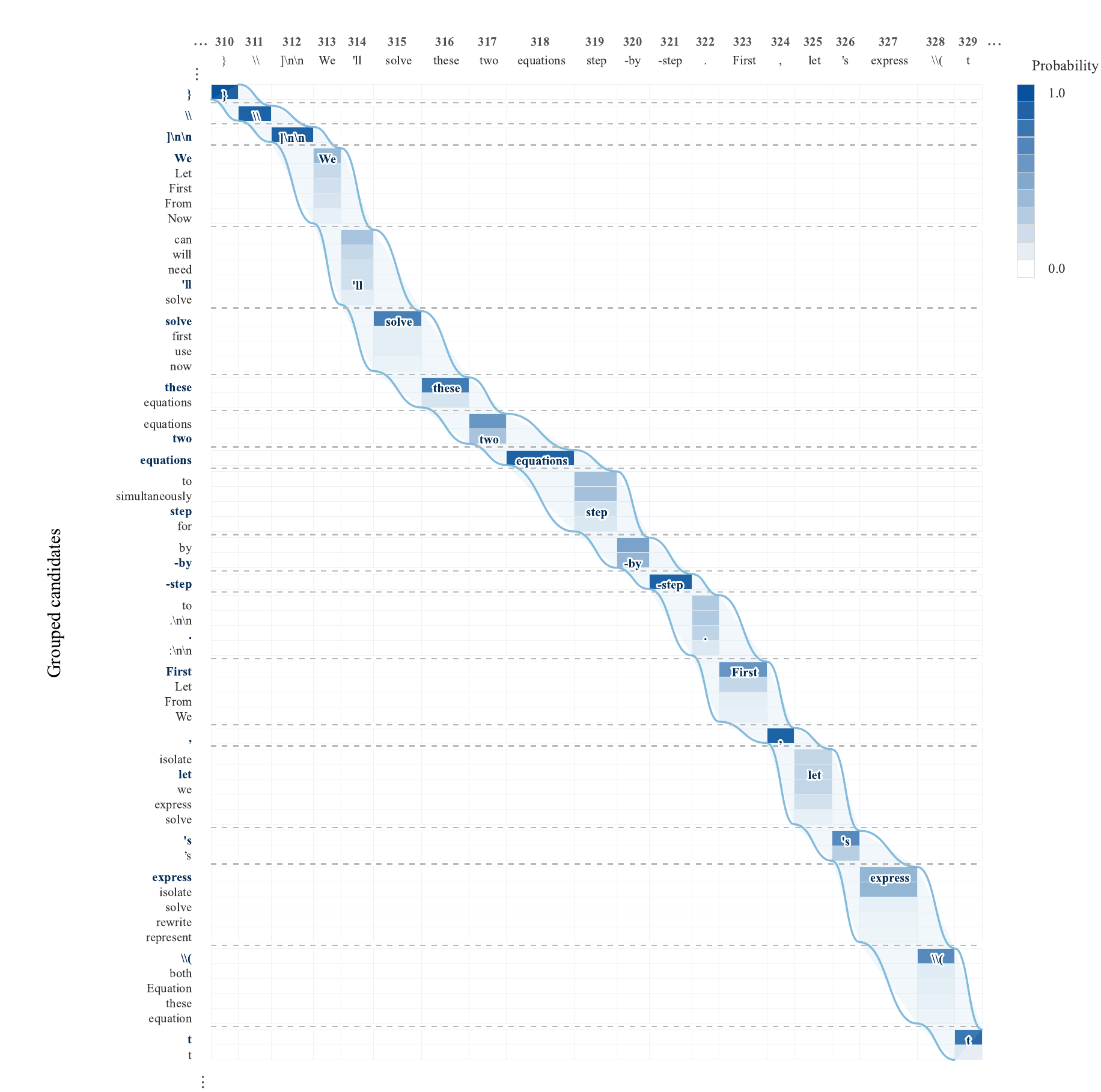}
\caption{Grouped candidate matrix for the CaSP-trained model on response tokens 310--329.
  CaSP preserves a clearer sampled-token path while maintaining non-negligible probability on nearby candidate alternatives, indicating less severe local distribution collapse.}
  \label{Qualitative_casp1}
\end{figure*}

\begin{figure*}[t!]
  \centering
  \setlength{\abovecaptionskip}{4pt}
  \setlength{\belowcaptionskip}{0pt}
  \includegraphics[width=0.99\textwidth]{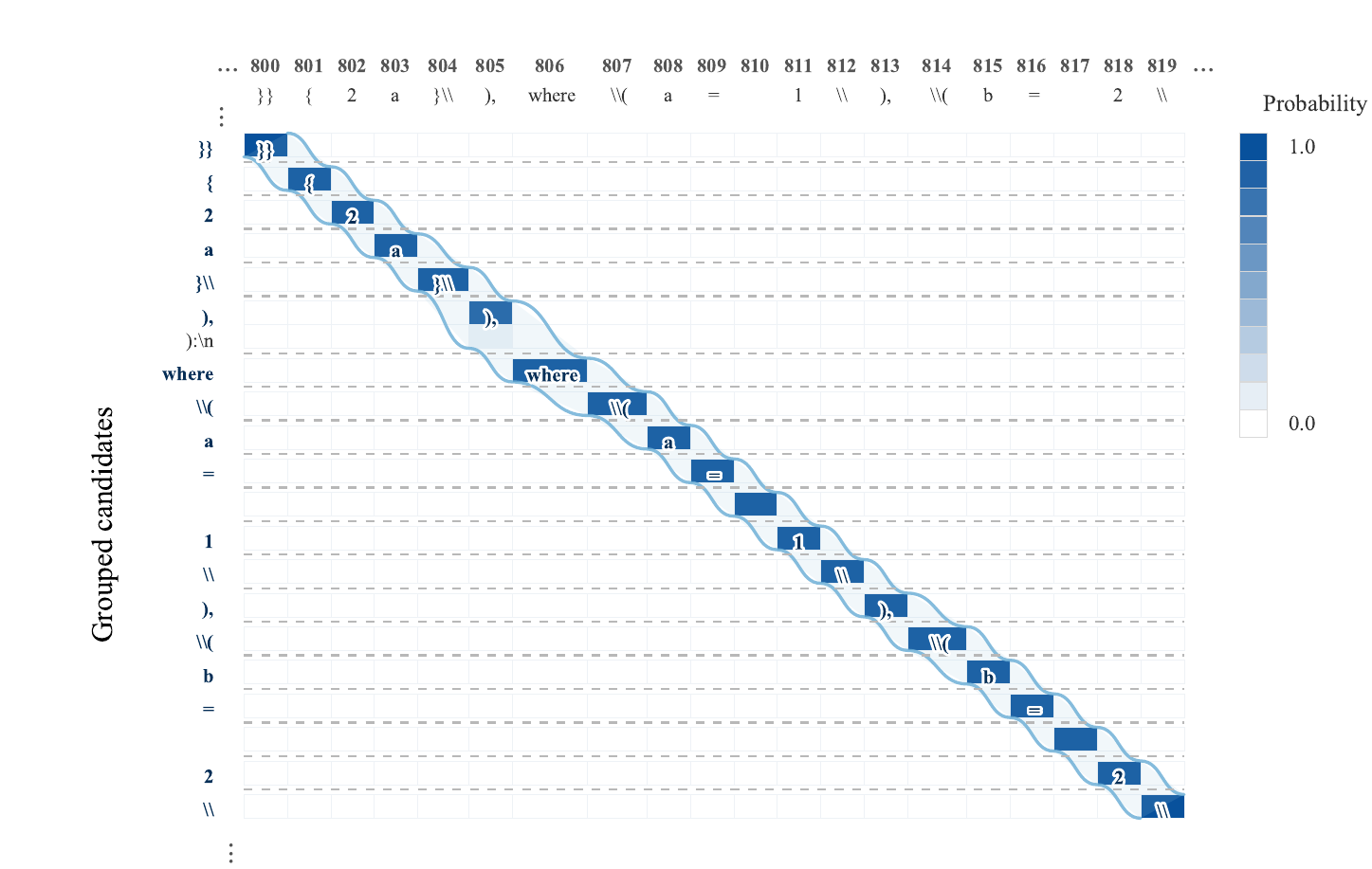}
  \caption{Grouped candidate matrix for the base model on response tokens 800--819.}
\label{Qualitative_base2}
\end{figure*}
\begin{figure*}[t!]
  \centering
  \setlength{\abovecaptionskip}{4pt}
  \setlength{\belowcaptionskip}{0pt}
  \includegraphics[width=0.99\textwidth]{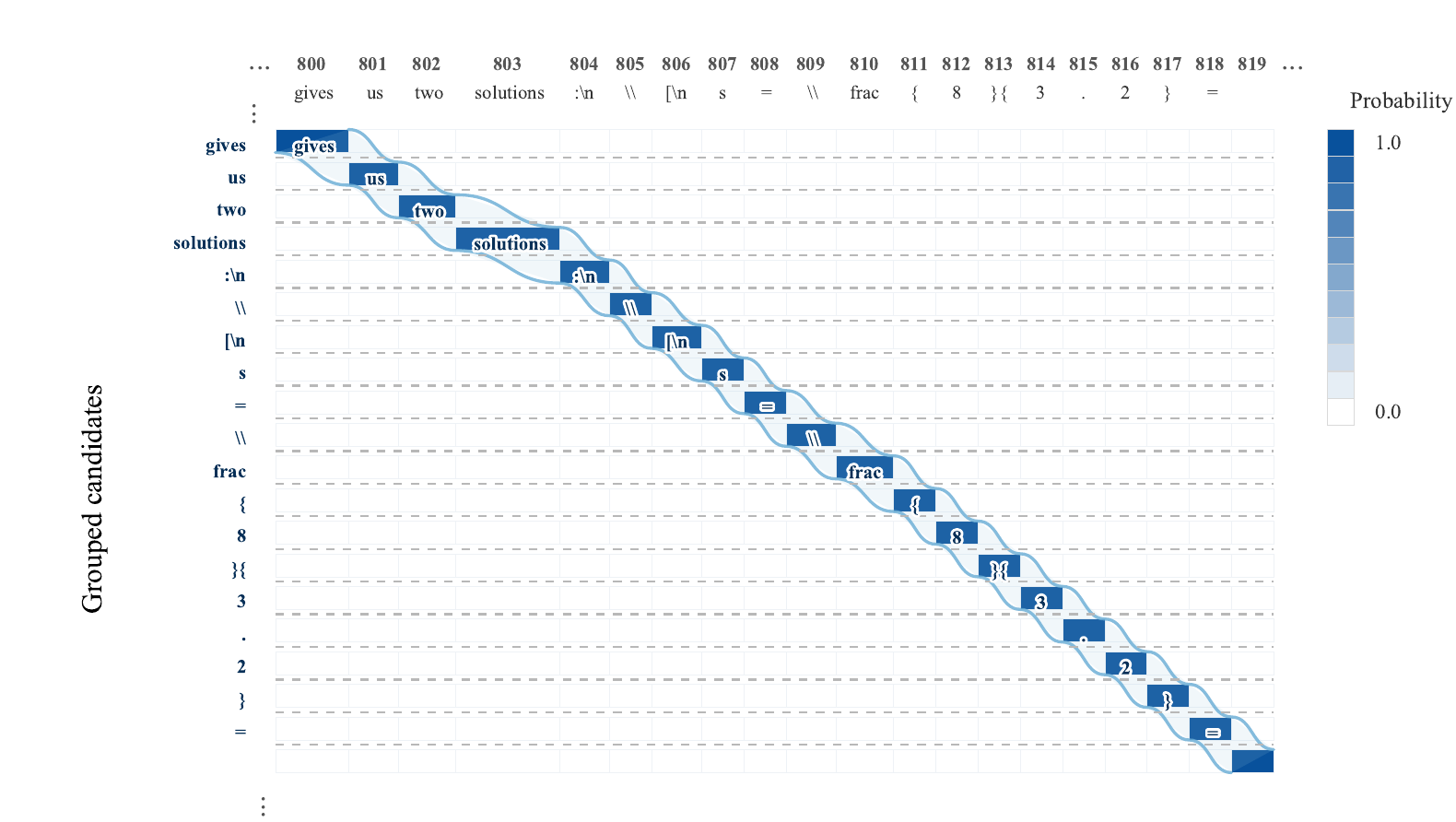}
  \caption{Grouped candidate matrix for the GRPO-trained model on response tokens 800--819.
  GRPO sharpens the local next-token distribution, concentrating probability on a narrow candidate path.}
\label{Qualitative_grpo2}
  
\end{figure*}
\begin{figure*}[t!]
  \centering
  \setlength{\abovecaptionskip}{4pt}
  \setlength{\belowcaptionskip}{0pt}
  \includegraphics[width=0.99\textwidth]{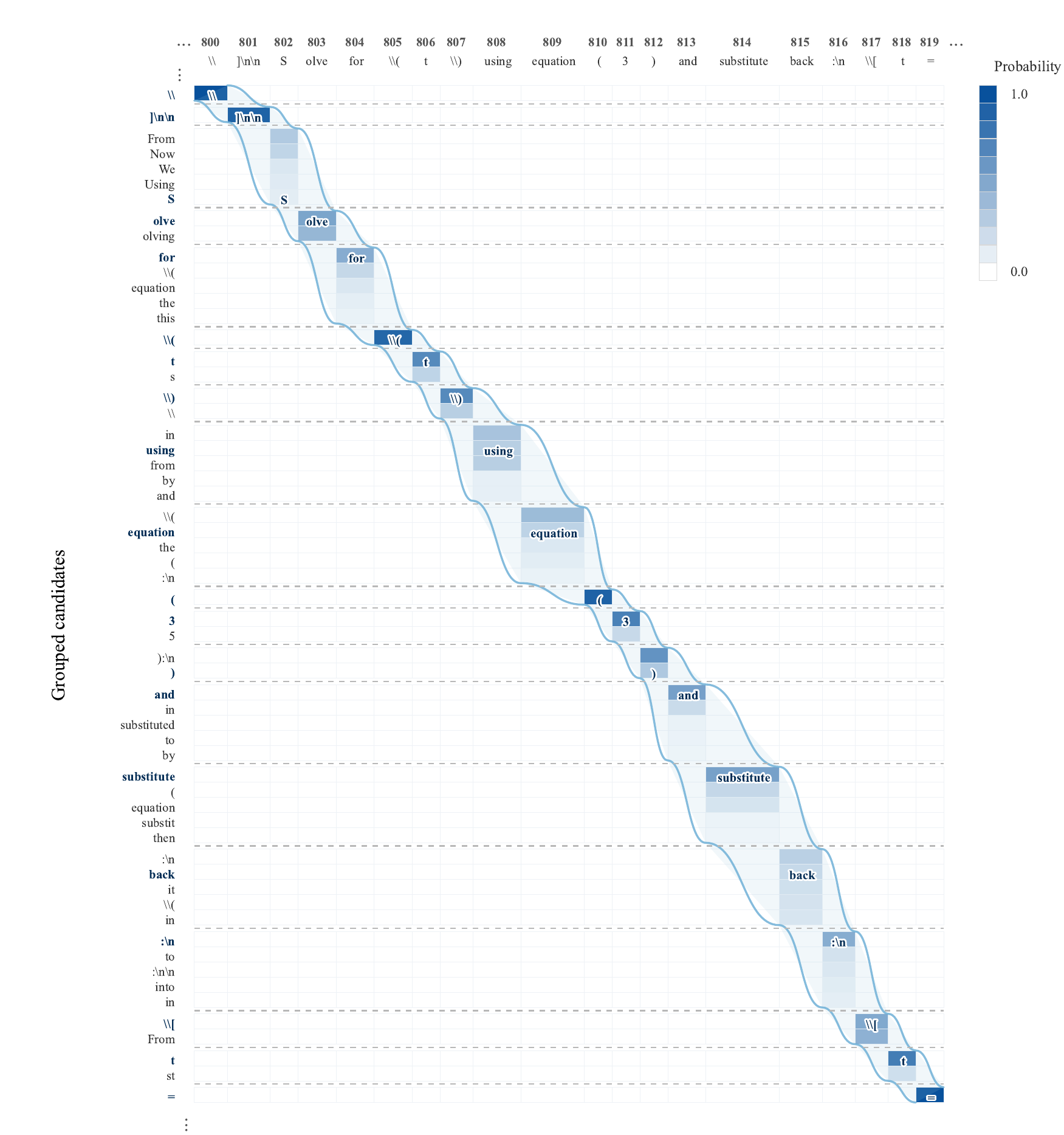}
\caption{Grouped candidate matrix for the CaSP-trained model on response tokens 800--819.
  CaSP maintains broader local candidate support than GRPO while preserving the sampled reasoning trajectory.}
  \label{Qualitative_casp2}
\end{figure*}

\end{document}